# The Artificial Intelligence Cognitive Examination: A Survey on the Evolution of Multimodal Evaluation From Recognition to Reasoning

**MAYANK RAVISHANKARA[1], (Member, IEEE), AND VARINDRA V. PERSAD MAHARAJ[2]**

[1]Independent Researcher, San Francisco, CA 94107, USA
[2]Independent Researcher, Sunnyvale, CA 94085, USA

Corresponding author: Mayank Ravishankara (mayankgowda@gmail.com)

**ABSTRACT** This survey paper chronicles the evolution of evaluation in multimodal artificial intelligence (AI), framing it as a progression of increasingly sophisticated "cognitive examinations." We argue that the field is undergoing a paradigm shift, moving from simple recognition tasks that test "what" a model sees, to complex reasoning benchmarks that probe "why" and "how" it understands. This evolution is driven by the saturation of older benchmarks, where high performance often masks fundamental weaknesses. We chart the journey from the foundational "knowledge tests" of the ImageNet era to the "applied logic and comprehension" exams such as GQA and Visual Commonsense Reasoning (VCR), which were designed specifically to diagnose systemic flaws such as shortcut learning and failures in compositional generalization. We then survey the current frontier of "expert-level integration" benchmarks (e.g., MMBench, SEED-Bench, MMMU) designed for today's powerful multimodal large language models (MLLMs). Next, we explore the uncharted territories of evaluating abstract, creative, and social intelligence. We also discuss the emerging "Post-2025" landscape, where rapid saturation of reasoning benchmarks by models such as Gemini 3 Pro necessitates a shift toward "living" adversarial evaluations. We conclude that the narrative of AI evaluation is not merely a history of datasets, but a continuous, adversarial process of designing better examinations that, in turn, redefine our goals for creating truly intelligent systems.

## I. INTRODUCTION: REDEFINING THE PASSING GRADE FOR AI

The rapid ascent of artificial intelligence (AI), particularly the proliferation of large language and multimodal models, has brought a foundational question into sharp focus: how do we measure progress? [1] This question is not merely academic; as AI systems are increasingly embedded in high-stakes domains from medicine to transportation, the methods used to evaluate their capabilities have become critical determinants of safety, reliability, and societal trust [2]. This survey paper presents a comprehensive analysis of the evolution of AI evaluation, proposing a narrative framework that views this history not as a simple timeline of datasets, but as a series of progressively more demanding *AI Cognitive Examinations* [3]. This metaphor casts the field's benchmarks as tests designed to probe ever-deeper levels of machine intelligence, moving from the equivalent of a vocabulary quiz to a graduate-level qualifying exam that assesses abstract, creative, and causal reasoning [4], [5], [6], [7], [8].

The primary catalyst for this evolution was the persistent *benchmark saturation* phenomenon. Repeatedly, the AI research community has developed what appears to be challenging tests, only to see them rapidly "solved" by the next generation of models. For instance, benchmarks such as BIG-Bench Hard (BBH), once the gold standard for reasoning, have been rendered effectively obsolete by models such as GPT-4o and Gemini, which now achieve zero-shot accuracy of 90.0% [9] and 84.0% [10] respectively, effectively confirming saturation. This outcome has necessitated the creation of benchmarks such as BIG-Bench Extra Hard (BBEH) [11] and highlights that these high-water marks are highly sensitive to prompt engineering and decoder strategies (e.g., Chain-of-Thought (CoT) vs. Direct Answer), a methodological consideration discussed in [12] and [13]. This recurring cycle exposes a critical flaw in static evaluation: a high score is not a reliable indicator of generalizable intelligence. It is a classic manifestation of Goodhart's Law: "when a measure becomes a target, it ceases to be a good measure" [13], [14]. Models are often "teaching to the test," optimizing benchmark-specific formats and exploiting statistical shortcuts rather than developing genuine cognitive abilities [13].

This realization has precipitated a fundamental shift in evaluation philosophy, a transition from asking *"what"* to asking *"why"* and *"how"* [15], [16]. Early benchmarks were concerned with what a model could identify—for example, "Is there a cat in this image?" [15]. The new frontier of evaluation is concerned with the process of inference itself: how a model integrates disparate pieces of information and why it arrives at a particular conclusion—for example, "Why does [person1] look surprised?" This mirrors the pedagogical shift in human education from emphasis on rote memorization to the cultivation of critical thinking and analytical skills. The goal is no longer just to get the right answer, but to get it for the right reasons.

### *A. TERMINOLOGY: MLLM VS. LVLM*

To ensure precision, we distinguish between two core terms used throughout this survey. *Multimodal Large Language Models (MLLMs)* is the overarching category for systems capable of processing diverse modalities (image, video, audio, text). *Large Vision-Language Models (LVLMs)* refers to the specific subclass of MLLMs tailored exclusively for image–text tasks (e.g., VQA, captioning). We use *LVLM* when discussing Level II static benchmarks and *MLLM* when addressing the broader, cross-modal capabilities of Level III and IV systems.

### *B. DISTINCTION FROM EXISTING BENCHMARKING SUITES*

Our survey is complementary to modern evaluation platforms and benchmark suites (e.g., HELM, OpenCompass) [1], [17], which provide standardized, multi-dimensional measurements of model performance across tasks and metrics. These toolkits are designed to answer questions of the form: "How well does model X perform on today's evaluation stack?"

In contrast, our contribution is not an additional benchmark or toolkit, but a pedagogical and historical framework for understanding why that evaluation stack looks the way it does. The *"Cognitive Examination"* traces how successive generations of benchmarks emerged in response to specific failure modes: for example, how limitations of static image recognition (Level I) [18] motivated the rise of visual question answering (Level II) [19], and how shortcut behaviors in those tasks [20] further pushed the community toward more reasoning-centric benchmarks (Level III) [21].

Rather than proposing new metrics or leaderboards, we systematize the evolutionary logic behind evaluation, linking concrete technical shifts (e.g., from recognition to compositional reasoning) to the pressures that made prior metrics inadequate. This historical, causal perspective is orthogonal to the operational focus of existing suites.

### *C. SCOPE AND LITERATURE CUTOFF*

This survey comprehensively covers the evolution of multimodal evaluation from the seminal ImageNet benchmark (2012) through to the current suite of cross-modal reasoning evaluations, with a specific focus on developments in 2023 and 2024. Our analysis includes literature published up to *October 1, 2025*. Any major benchmark or architectural development published after this date (e.g., Gemini 3, OpenAI o3) is referenced in the *Future Outlook (Section IX-B)* to maintain a clear and consistent scope for the survey. This demarcation ensures that the historical framework remains coherent while acknowledging emerging, potentially unpublished work.

### *D. SURVEY ROADMAP: THE FOUR-LEVEL COGNITIVE FRAMEWORK*

This survey is structured into a hierarchy of four progressively demanding AI Cognitive Examinations, designed to navigate the historical evolution of evaluation standards:

- *Level I: Foundational Knowledge (The Recognition Era)* will explore landmark benchmarks such as ImageNet and Common Objects in Context (COCO) which established the field's core "knowledge base" for object recognition and scene description.
- *Level II: Applied Logic and Comprehension (The Dawn of Reasoning)* will detail the creation of diagnostic benchmarks such as GQA and Visual Commonsense Reasoning (VCR), which were explicitly designed to probe and expose systemic flaws—such as shortcut learning and poor compositional generalization—that were masked by high scores on earlier tests.
- *Level III: Expert-Level Multimodal Integration (Current Frontier)* will analyze the new wave of holistic benchmarks (MMBench, SEED-Bench, MMMU) created to evaluate the complex, cross-modal reasoning capabilities of today's powerful multimodal large language models (MLLMs) [22]. Crucially, we critique the reliance on "outcome-based" metrics (accuracy) in this era, advocating for the "process-based" fidelity (Chain-of-Thought scoring) that defines the next frontier.
- *Level IV: Abstract and Creative Intelligence (Uncharted Territories)* will venture into the emerging and challenging domains of evaluating embodied planning, social intelligence, and creativity, where objective right-or-wrong answers give way to more subjective and behavioral assessments.

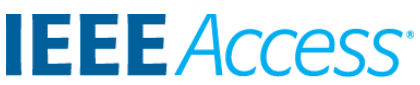

By tracing this path, this study argues that the history of AI evaluation is not a passive chronicle of datasets. Rather, it is an active, often adversarial, diagnostic cycle. Researchers, acting as cognitive scientists, have designed new examinations to falsify the hypothesis of machine intelligence presented by the previous generation of models. In doing so, they not only measure progress but also actively redefined the goals of the field, pushing the frontier of perception beyond simple recognition and toward true multimodal reasoning.

## II. THE MULTIMODAL MEASUREMENT FRAMEWORK

While the Cognitive Examination described in this survey follows a historical narrative, it is necessary to ground these evolutionary stages in a rigorous measurement framework. Critics of early AI evaluation often point to a conflation of task difficulty, modality, and evaluation method. To address this, we present the *Multimodal Measurement Framework* (Table 1), which disentangles these dimensions across the four levels of cognitive evolution.

This framework evaluates benchmarks along six orthogonal axes:

- *Skill:* The specific cognitive capability being tested (e.g., Pattern Matching vs. Reasoning).
- *Reliability:* The statistical consistency of the metric (e.g., Top-1 Accuracy vs. Elo Rating).
- *Robustness:* Mechanisms included to prevent shortcut learning (e.g., Out-of-Distribution sets).
- *Hygiene:* Measures taken to ensure data validity and prevent contamination.
- *Cost:* The computational resource required for evaluation (e.g., Static Inference vs. Simulation).
- *Fairness:* Mechanisms to detect bias and ensure representation across diverse groups.

By viewing the benchmarks through this lens, we can see that the transition from Level I to Level IV involves trade-offs: while *Skill* depth increases, *Cost* rises exponentially, and *Reliability* often decreases as metrics shift from deterministic accuracy to stochastic judgment.

### A. THE AUDIT RECIPE: DATA CONTAMINATION

Across the four levels, we observe four distinct "audit recipes" for data contamination and benchmark hygiene. Each era emphasizes a different mechanism for keeping evaluation sets meaningful as models and training data change.

*Level I – Structural Isolation.* Early perception benchmarks rely on *structural isolation*: fixed train/validation/test splits, withheld test labels, and centralized evaluation servers (e.g., ImageNet/ILSVRC [18], [23], PASCAL VOC [24], MS COCO [25]). Contamination is framed mainly as train–test leakage within the benchmark, addressed by hidden test sets and challenge servers [23].

*Level II – Structural Isolation + diagnostic stress-tests.* Level II benchmarks keep the same static-split recipe (e.g., VQA [19], GQA [26], NLVR2 [27], VCR [28]) but add diagnostic stress-tests for shortcut behaviour and dataset bias, such as VQA-CP's re-splitting by answer distribution [29] and CLEVR/NLVR's controlled constructions [30], [31]. These diagnostics focus on shortcuts rather than web-scale pre-training leakage, so hygiene still rests on structurally isolated yet public static test sets.

*Level III – Post-hoc diagnostic audits.* Expert multimodal benchmarks (e.g., MathVista [32], MMMU [21], MMBench [17], MM-Vet [33], HallusionBench [34], Video-MME [35]) again use static structural isolation, but now pre-training data are web-scale and opaque, and many items come from exams, textbooks, or widely available web content [2]. The ecosystem therefore leans on post-hoc diagnostic audits (n-gram overlap [36], [37], IR-based overlap [2], TS-Guessing [38]) to detect contamination, shifting from "we assume the test set is clean" to "we suspect it is dirty and try to measure how bad the leakage is."

*Level IV – Dynamic Prevention.* Level IV benchmarks adopt dynamic prevention: human-in-the-loop platforms (e.g., Dynabench [13]) that continually collect new adversarial examples, time-sensitive QA (e.g., RealTimeQA [39] and later TSQA work such as FreshQA [40], TAQA [41]) that builds test data after a model's cutoff, and embodied challenges (e.g., Habitat [42], VirtualHome [43], BEHAVIOR [44]) that evaluate agents in unseen environments. Together, HITL data collection and temporal or environmental sequestering keep the test distribution in front of the training data and make memorization of a fixed benchmark increasingly unproductive.

### B. INTERPRETING HYGIENE SCORES ACROSS LEVELS

At the end of each level, we assign a qualitative hygiene score summarizing how well that level's dominant audit recipe guards against contamination and benchmark overfitting. These scores enable rough "apples-to-apples" comparisons across levels (Structural Isolation in Levels I–II, post-hoc diagnostic audits in Level III, and Dynamic Prevention in Level IV). Because these recipes differ in kind, projecting them onto a single 1–5 scale necessarily involves some subjective judgment; the scores are best read as a scientifically informed heuristic, grounded in concrete criteria such as test-set publicity, evidence of train–test overlap, and the presence of code-upload or temporal sequestering.

## III. LEVEL I: FOUNDATIONAL KNOWLEDGE (THE RECOGNITION ERA)

The modern era of AI is built upon the foundation of a large-scale, standardized test that focuses on core cognitive skills: recognition. These benchmarks provided the common ground upon which the deep learning revolution was built, establishing a shared vocabulary of tasks, metrics, and objectives. These were the field's first comprehensive "knowledge tests,'' designed to answer the fundamental question: Can a machine learn to see the world as we do?

**TABLE 1. The multimodal measurement framework: Mapping cognitive levels to rigorous evaluation dimensions.**

| Level | *Skill* | *Reliability* | *Robustness* | *Hygiene* | *Cost* | *Fairness* |
|---|---|---|---|---|---|---|
| *I. Perception* | Pattern Matching & Feature Extraction | Top-1 / Top-5 Accuracy, mAP | ImageNet-C (Texture/Noise variants) | Fixed Train/Test splits | Low (Static Inference) | Geo-diversity checks (e.g., Dollar Street) |
| *II. Alignment* | Visual Grounding & Semantic Binding | BLEU, METEOR, SPICE (N-gram) | Balanced-VQA (Prior de-biasing) | Human-verified test sets | Moderate (Visual Enc.) | Social Bias probes (e.g., VQA-CP) |
| *III. Reasoning* | Multi-step Logic, Temporal Causality | Exact Match (EM), CoT Accuracy | Instruction Following (If-Then) | Dynamic/Private Test sets (MMBench) | High (Large Language Model (LLM) Generation) | Safety & Toxicity Red-teaming |
| *IV. Embodiment* | Long-horizon Planning & Creativity | LLM-as-a-Judge, Elo Ratings, SR | Adversarial Prompting (Jailbreaks) | Canary Strings & Temporal Cutoffs | Very High (Physics Sim) | Interaction & Stereotype bias |

The Recognition Era, approximately 2009–2015, began with the release of ImageNet [18], reached its pivotal month with AlexNet's 2012 breakthrough [45], and was concluded as the ImageNet Large Scale Visual Recognition Challenge (ILSVRC) retrospective [23] and the introduction of COCO [25] shifted evaluation beyond classification.

### A. *ImageNet/ILSVRC: STANDARDIZING THE RECOGNITION "EXAM"*

#### 1) WHAT IT IS

ImageNet [18] defined the modern era of large-scale visual recognition by introducing a WordNet-organized database that ultimately reached over 14 million hand-annotated images across more than 20,000 categories (''synsets'') [18], [23]. ImageNet project supplied the scale and diversity missing from earlier computer-vision datasets [18]. The annual ImageNet Large Scale Visual Recognition Challenge (ILSVRC, started in 2010) provided a competitive benchmark that popularized Top-1 and Top-5 accuracy for image classification [2], [16], [23].

#### 2) COVERAGE AND DESIGN

ILSVRC standardized the evaluation on a 1,000-class subset of ImageNet with fixed train/val/test splits and a public leaderboard, setting clear tasks and metrics for direct comparison across models [2], [23]. Models are evaluated by their ability to assign the correct class label to each image. The Top-5 metric counts a prediction as correct if the true label is among the five most confident outputs [2], [16]. This format became the field's canonical ''closed-book exam'' for recognition-focused AI.

#### 3) WHAT MEASURES (AND CATALYZES)?

By emphasizing large-scale supervised classification with simple, comparable metrics, ImageNet/ILSVRC measured a model's capacity for high-accuracy visual recognition at scale and catalyzed rapid innovation in architectures, training procedures, and computation. It serves as a shared yardstick for scaling laws and representation quality in the Recognition Era [2], [23].

**TABLE 2. ImageNet/ILSVRC benchmark at a glance.**

| Property | Value |
|---|---|
| Modalities | Natural images (single-label classification) |
| Design | ILSVRC: 1,000-class subset of ImageNet; fixed splits; public leaderboard [2], [23] |
| Primary metrics | Top-1 and Top-5 accuracy [2], [16] |
| Diagnostic focus | Scalable recognition performance; architecture and training innovations; field-wide comparability [2], [23] |
| Example milestone | AlexNet (2012): 15.3% Top-5 error; −10.8 percentage points (pp) vs. runner-up [23], [45] |
| Known limitations | Shortcut learning, texture bias, Out-of-Distribution (OOD) brittleness [46]–[49]; dataset collection biases [2], [23] |

#### 4) REPRESENTATIVE FINDINGS

The pivotal 2012 breakthrough by AlexNet achieved a Top-5 error of 15.3%—a 10.8 percentage-point improvement over the next best entry, demonstrating the power of deep convolutional networks and igniting the modern deep-learning revolution [23], [45]. Sustained leaderboard progress through 2017 established ImageNet as the benchmark record for perception-level AI [2], [23].

#### 5) WHAT IT LACKS (AS AN EVALUATION)

Despite its saturated performance, ImageNet is vulnerable to real-world generalization: models often rely on *shortcut* or non-robust features, displaying strong texture bias and poor out-of-distribution robustness [46], [47], [48], [49]. Crowdsourced collection pipelines and cost-efficient annotation introduced subtle, systematic biases that created iconic, single-object views overrepresented—conditions under which high accuracy can be achieved by exploiting dataset regularities rather than learning causal, semantically meaningful features [2], [23], [48]. Later stress tests such as ImageNet-C and ObjectNet further highlighted these weaknesses, motivating the move beyond pure classification toward richer evaluations at the next level.

As summarized in Table 2, ImageNet/ILSVRC standardized large-scale recognition with fixed splits and Top-1/Top-5 accuracy.

### B. *PASCAL VOC: SETTING THE RULES FOR DETECTION AND SEGMENTATION (2005–2012)*

#### 1) WHAT IT IS

The PASCAL Visual Object Classes (VOC) Challenge established a shared benchmark and evaluation protocol for visual recognition, spanning image classification, object detection, semantic segmentation, person layout, and action classification across 20 categories [24], [50]. From 2005 to 2012, VOC's fixed train/val/test splits and centralized evaluation server enabled rigorous, repeatable comparisons and a stable leaderboard culture that shaped community practice [24], [50].

**TABLE 3. PASCAL VOC benchmark at a glance.**

| Property | Value |
|---|---|
| Modalities | Natural images; tasks: classification, detection, segmentation, actions, person layout |
| Design | Fixed splits + evaluation server; 20 object categories; yearly challenge (2005–2012) [24], [50] |
| Primary metrics | Detection: AP/mAP at IoU $\geq$ 0.5; Segmentation: mean IoU [24] |
| Diagnostic focus | Localization quality (IoU), precision–recall behavior, multitask robustness [24], [50] |
| Example milestone | Community-wide standardization of AP/mAP and IoU; influence on subsequent large-scale benchmarks [50] |
| Known limitations | Small label space; saturation at IoU@0.5; limited long-tail coverage [50] |

#### 2) COVERAGE AND DESIGN

VOC crystallized the now-standard detection protocol: Average Precision (AP) and mean AP (mAP) computed under the *PASCAL overlap criterion*—an intersection-over-union (IoU) threshold of $\geq$ 0.5 between predicted and ground-truth boxes [24]. Semantic segmentation was evaluated using mean intersection-over-union (mIoU) across classes, and annotations included *difficult* and *truncated* flags to facilitate nuanced error analysis [24], [50]. Annual editions (2005–2012) held categories and rules steady while evolving tasks and diagnostics, allowing progress to be tracked cleanly over time [50].

#### 3) WHAT IT MEASURES (BEYOND ACCURACY)

By tying localization quality (IoU) to precision–recall–derived AP, VOC evaluated not only whether a model recognized an object but also *how well it localized it*. This unified recognition and localization under a single, comparable protocol and encouraged methods that balance precision and recall rather than optimizing a single accuracy figure [24], [50].

#### 4) REPRESENTATIVE FINDINGS

Across the VOC years, results document steady gains from pre–deep learning pipelines (e.g., hand-crafted features and part-based models) toward learned, end-to-end approaches. Retrospectives highlight that iterative improvements in features, part modeling, and later convolutional architectures translated into consistent progress on the same splits, helping standardize AP/mAP and IoU as community-wide metrics and influencing subsequent large-scale benchmarks [50].

#### 5) WHAT IT LACKS (AS AN EVALUATION)

The benchmark's modest label space (20 categories) and relatively iconic views limited stress on long-tail variation and ultimately led to metric saturation. Its single-threshold IoU@0.5 emphasizes coarse localization fidelity; later benchmarks broadened category breadth and difficulty and adopted evaluation over multiple IoU thresholds to address VOC-era ceiling effects [50].

Key properties of the PASCAL VOC benchmark, including tasks, metrics, and protocol are outlined in Table 3.

### C. COCO: CONTEXTUAL, MULTI-OBJECT PERCEPTION (2014–)

#### 1) WHAT IT IS

The Common Objects in Context (COCO) dataset reframed recognition as understanding scenes with *multiple* interacting objects in their natural environments [25]. Unlike single-label classification, COCO spans detection, instance segmentation, keypoint detection (human pose), and image captioning, pushing evaluation toward richer, context-sensitive perception [2], [25], [51].

#### 2) COVERAGE AND DESIGN

COCO contains complex, everyday scenes with many objects, frequent occlusion, and clutter—the antithesis of iconic views emphasized in early classification datasets [25]. Annotations support *object detection* (1.5M+ instances across 80 categories), *instance segmentation* (pixel-accurate masks), *keypoint detection* (17 human keypoints), and *captioning* (multiple human-authored captions per image) [2], [51]. In addition to "thing" categories (countable objects), COCO introduced a taxonomy for "stuff" (amorphous regions such as sky, grass, road), strengthening the role of background and context in evaluation [25].

#### 3) WHAT IT MEASURES (BEYOND ACCURACY)

COCO formalized more discriminative metrics than earlier detection benchmarks: mean Average Precision (mAP) *averaged across IoU thresholds* from 0.50 to 0.95 (in 0.05 steps), alongside $AP_{50}$, $AP_{75}$, and size-specific breakdowns ($AP_S$/$AP_M$/$AP_L$). The same AP protocol applies to masks for instance segmentation; keypoint detection uses a keypoint-similarity variant. Captioning tasks evaluate semantic adequacy and fluency using established automatic metrics. Together, these protocols measure not only "is it there?" but also *how precisely it is localized*, *how well instances are separated*, and *how much context the model captures* [2], [25], [25].

#### 4) REPRESENTATIVE FINDINGS

COCO's harder scenes and stricter metrics became the proving ground for high-precision localization and dense prediction, accelerating advances in detectors and instance segmentation systems and becoming the default benchmark for real-time detectors as well [25]. Yearly challenges and public leaderboards entrenched mAP@[.50:.95] as the community's currency for progress, with strong empirical gaps between $AP_{50}$ and $AP_{75}$/$AP_{.50:.95}$ highlighting the need for precise localization and robust instance delineation [2], [25].

**TABLE 4. COCO benchmark at a glance (Level I—Recognition).**

| Property | Value |
|---|---|
| Modalities & tasks | Natural images; detection, instance segmentation, keypoint detection (pose), image captioning [25], [51] |
| Design | Complex multi-object scenes with occlusion/clutter; "things" + "stuff" taxonomy; fixed splits [25] |
| Primary metrics | mAP averaged over IoU 0.50:0.95 (and $AP_{50}$, $AP_{75}$); size-wise $AP_{S/M/L}$; mask AP for instances; standard caption metrics [2], [25]. |
| Diagnostic focus | Precise localization, instance separation, context sensitivity; pose estimation; caption grounding [2], [25] |
| Example milestone | Establishment of mAP@[.50:.95] and dense prediction tasks as field standards; catalyst for modern detection/segmentation pipelines [2], [25] |
| Known limitations | Static images; fixed categories; limited long-tail/cause; eventual saturation at the top end [25] |

#### 5) WHAT IT LACKS (AS AN EVALUATION)

Despite its breadth, COCO focuses on *static* images and object-centric tasks. It offers limited coverage of long-tail categories, causal reasoning, or cross-modal integration, and its fixed categories and scenes can saturate for top systems. These gaps motivated extensions emphasizing "stuff" coverage, longer-tailed distributions, video/temporal context, and tasks beyond perception (e.g., compositional reasoning addressed in higher levels of our hierarchy) [2], [25].

COCO's tasks, metrics, and design choices are summarized in Table 4.

### D. CRACKS IN THE FOUNDATION: EARLY SIGNS OF BRITTLENESS

Despite progress driven by ImageNet and COCO, a growing body of research has revealed that high performance on these benchmarks is a fragile indicator of real-world competence. The benchmarks, it turned out, were only proxies for the true task of visual understanding, and success on the proxy did not always transfer to practical applications where data distributions differ [52].

The source of this brittleness can be traced back to the datasets. For example, the crowdsourced data-collection pipeline for ImageNet has an inherent bias. Annotators on Mechanical Turk were typically shown an image retrieved for a specific class (e.g., 'airliner') and asked to confirm if the image contained an object of that class [48]. This process is framed as a leading question and is naturally filtered for images with clear, unambiguous, and often iconic views of the target object. Consequently, the final dataset was heavily skewed toward single-object images, with one study finding that models suffered a 10% accuracy drop on multi-object images compared with single-object images [48].

More insidiously, this process creates strong, spurious correlations between objects and their contexts. The models learned to exploit these statistical regularities as shortcuts. For instance, a model might achieve high accuracy on the class "pickelhaube" (a type of spiked helmet) not by learning the features of the helmet itself, but by recognizing the co-occurring, and often more salient, "military uniform" [48]. This strategy, although effective for improving ImageNet accuracy, fails to generalize to real-world scenarios where a pickelhaube might appear without a uniform [48]. These findings revealed a deep-seated problem: the models did not learn the essence of objects, but instead learnt to exploit the statistical quirks of the dataset. In many cases, high scores were an illusion of competence built on the foundation of brittle, non-causal correlations [13]. This realization marked the end of the first era of evaluation and set the stage for a new generation of benchmarks designed not only to measure performance but also to diagnose these fundamental flaws.

### E. SYNTHESIS OF LEVEL I: THE PERCEPTION ERA

The benchmarks of the Perception Era (ImageNet [18], PASCAL VOC [24], COCO [25]) fundamentally established the scientific standard for computer vision, prioritizing statistical consistency above all else. As synthesized in Table 5, this era achieved the highest possible score on *Reliability* due to the deterministic nature of its metrics (Top-1 Accuracy, mAP) and the use of static, fixed test sets [23]. *Cost* efficiency was also optimal, as evaluation required only discriminative classification rather than generative inference.

However, the *Robustness* and *Fairness* axes reveal the era's critical flaws. As detailed in the "Cracks in the Foundation" subsection, high reliability scores often masked a reliance on brittle texture shortcuts (low Robustness) [47] and Western-centric data distributions (low Fairness) [48]. The transition to Level II was driven not just by a desire for harder tasks, but by the specific need to fix these broken axes.

## IV. LEVEL II: APPLIED LOGIC AND COMPREHENSION (THE DAWN OF REASONING)

While Level I benchmarks primarily assess fundamental pattern recognition and perception, Level II focuses on *applied logic and comprehension*,testing a model's ability to integrate perception with reasoning, multi-step inference, and natural language understanding. Benchmarks at this level move beyond simple recognition to evaluate whether models can connect visual and textual cues, apply compositional reasoning, and avoid reliance on dataset-specific statistical shortcuts [2], [20], [53]. This stage represents a pivotal transition: success requires not only accurate perception but also a deeper understanding of relationships, causal links, and contextual meaning. Historically, many multimodal systems have achieved strong surface performance yet fail under distribution shifts or adversarial perturbations, motivating increasingly sophisticated evaluation benchmarks [13], [54], [55].

As researchers have observed high-performing models exploiting statistical biases rather than truly understanding inputs, a new class of benchmarks has emerged. These were designed not only to grade the final answer, but also to explicitly probe reasoning, compositionality, and commonsense—forcing models to demonstrate not only *what* they see, but also *how* they think. This marked the beginning of the *Reasoning Era* (approximately 2015–2020), led by seminal datasets such as visual question answering (VQA) [19] and GQA [26]. These benchmarks moved beyond recognition toward compositional reasoning and diagnostic evaluation, exposing the brittleness and superficial pattern-matching behavior of recognition-era systems [52].

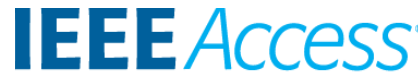

**TABLE 5.** Framework synthesis: Level I (Perception).

| Dimension | Rating | Justification and Evidence |
|---|---|---|
| *Skill* | ⋆⋆ | *Foundational.* Limited to pattern recognition and localization (ImageNet [18], COCO [25]). While sufficient for identifying objects, it lacks the logical, causal, or compositional reasoning required for higher-level cognition. |
| *Reliability* | ⋆⋆⋆⋆⋆ | *Gold Standard.* Deterministic metrics (Top-1 Accuracy, mAP) allow perfect reproducibility across labs. Evaluation servers (ILSVRC [23], PASCAL [25]) ensured consistent scoring protocols. |
| *Robustness* | ⋆ | *Critical Failure.* Models exploited texture shortcuts rather than shape [47] and failed on distribution shifts (ImageNet-C [4]), revealing that high accuracy did not imply robust vision. |
| *Hygiene* | ⋆⋆⋆⋆ | *High.* Strict adherence to fixed Train/Val/Test splits and centralized evaluation servers prevented the massive data leakage seen in modern web-scraped training. |
| *Cost* | ⋆⋆⋆⋆⋆ | *Optimal.* Static discriminative inference (e.g., ResNet forward pass) is computationally cheap and highly scalable compared to generative autoregression. |
| *Fairness* | ⋆ | *Critical Failure.* Severe geographic and social bias in web-scraped collection. Analysis shows models suffer accuracy drops on non-Western images due to dataset skew [48]. |

The following sections trace this evolution, beginning with VQA and progressing toward increasingly challenging datasets that test deeper forms of multimodal understanding, commonsense reasoning, and robustness.

### *A. VISUAL QUESTION ANSWERING (VQA)*

#### 1) WHAT IT IS

The Visual Question Answering (VQA) benchmark [19] was a pioneering effort to unify computer vision and natural language processing by requiring systems to answer open-ended, natural-language questions about images. Given an image and free-form question (e.g., "What color are the cat's eyes?"), the models must generate a concise answer. This benchmark has quickly become one of the most widely adopted tasks for evaluating multimodal understanding, spurring rapid advances in both architectures (e.g., attention mechanisms and transformers) and training paradigms.

#### 2) COVERAGE AND DESIGN

The original VQA dataset [19] consisted of over 200,000 images sourced from MS-COCO and abstract scenes, paired with more than 760,000 human-authored questions across diverse categories such as object identification, counting, activity recognition, and commonsense reasoning. Each question was paired with ten independent human-provided answers to capture linguistic variability and ambiguity. The performance was evaluated using a consensus-based soft accuracy metric:

$$\text{Accuracy} = \min\left(\frac{\#\text{humans that gave answer}}{3}, 1\right)$$

This scoring method rewards partial agreement among annotators, mitigating the issues caused by inherently ambiguous or open-ended questions.

Building on this foundation, VQA v2.0 [56] expanded the dataset to approximately 1.1M questions across 204k images, explicitly balancing question-answer pairs to reduce language priors and force stronger reliance on visual grounding. Further extensions introduced specialized evaluation settings: VizWiz [57] focused on real-world accessibility with images taken by limited vision photographers, whereas Outside Knowledge Visual Question Answering (OK-VQA) [58] and A-Outside Knowledge Visual Question Answering (OKVQA) [59] evaluated knowledge-grounded reasoning that integrates external world knowledge.

#### 3) WHAT IT MEASURES (BEYOND ACCURACY)

VQA evaluates several intertwined capabilities: (1) grounding of natural language in visual content, (2) basic logical and spatial reasoning, (3) multi-step compositional reasoning such as counting and comparing attributes, and (4) integration of perception with background knowledge. However, strong aggregate performance can mask shallow reasoning strategies. For instance, early models exploited statistical shortcuts such as answering "yes" to most yes/no questions or guessing the most frequent color [20], [29]. These weaknesses motivated diagnostic variants such as VQA under changing priorities (VQA-CP) [29], which tests a model's ability to handle distributional shifts in question-answer priors, and GQA-out-of-distribution (OOD) [60], which probes out-of-distribution compositional reasoning.

#### 4) REPRESENTATIVE FINDINGS

(1) Attention-based models such as the Stacked Attention Network (SAN) [61] and Bilinear Attention Network (BAN) [62] initially dominated performance, but were later surpassed by transformer-based architectures such as LXMERT (Learning Cross-Modality Encoder Representations from Transformers (LXMERT) [63], which integrates visual and linguistic streams through cross-modal attention, and ViLBERT (Vision-and-Language BERT) [64], which extends BERT with dual attention pathways for image-text fusion. (2) Despite the high scores on standard splits, studies have revealed that performance improvements often stem from dataset scale and language modeling advances rather than true visual reasoning, as shown by adversarial perturbation studies [53]. (3) Even state-of-the-art models experience severe accuracy drops under domain shift or compositional evaluations, highlighting the persistent gap between superficial pattern recognition and robust multimodal comprehension [60].

**TABLE 6. VQA benchmark at a glance.**

| Property | Value |
|---|---|
| Modalities | Image + Text (questions and answers) |
| Scale | VQA v1: 200k images, 760k questions; VQA v2: 204k images, 1.1M questions |
| Tasks | Open-ended and multiple-choice visual question answering |
| Primary metric | Consensus-based soft accuracy |
| Diagnostic focus | Visual grounding, compositional reasoning, bias detection |
| Successor benchmarks | VQA-CP, GQA, OK-VQA, A-OKVQA |

**TABLE 7. VQA-CP benchmark at a glance.**

| Property | Value |
|---|---|
| Modalities | Natural images + text (questions and answers) |
| Design | Identical images/questions as VQA v2, but train/test splits swap frequent and rare question–answer pairs |
| Primary metric | Consensus-based soft accuracy (same as VQA) |
| Diagnostic focus | Robustness to distribution shifts, bias reduction, visual grounding |
| Example baseline | GRAD [68], RUBi [69], HINT [67] |
| Known limitation | Tests only language-prior shift, not full spectrum of OOD generalization |

#### 5) WHAT IT LACKS (AS AN EVALUATION)

VQA's open-ended format of VQA poses intrinsic evaluation challenges: consensus-based soft accuracy cannot fully capture nuanced reasoning, multi-step inference, or partial correctness. Moreover, the dataset's naturalistic collection process introduces strong annotations and language priors, enabling shortcut exploitation rather than genuine visual grounding [20]. Consequently, VQA has evolved into a foundational, entry-level benchmark that primarily tests surface-level integration of vision and language. It inspired the development of specialized successor benchmarks — VQA-CP for bias analysis, GQA for compositional reasoning, and OK-VQA/A-OKVQA for knowledge-grounded reasoning — which together form a more comprehensive evaluation landscape.

Key properties of the VQA benchmark, including splits and metrics, are listed in Table 6.

### B. VQA-CP: TESTING ROBUSTNESS TO BIAS AND SHORTCUT LEARNING

#### 1) WHAT IT IS

Although the original VQA benchmark [19] catalyzed multimodal research, subsequent analyses revealed that many models achieved high scores by exploiting *language priors* rather than genuine visual grounding [20], [65]. For example, if most questions beginning with "Is the banana…" were answered "yes" in the training set, models learned to default to "yes" without attending to the image. The VQA Under Changing Priors benchmark (VQA-CP) [29] was introduced to explicitly diagnose this failure mode by *deliberately altering the distribution of question–answer pairs between training and testing*.

#### 2) COVERAGE AND DESIGN

VQA-CP is derived from the same underlying images and questions as VQA v2 [56], but reorganizes the train and test splits such that frequent question–answer pairs in training are *rare* at test time, and vice versa. For instance, if the training data contained 80% yellow bananas and 20% non-yellow bananas, the test data flipped this to 20% yellow and 80% non-yellow bananas. This design forces the models to *rely on visual evidence* instead of memorizing language-based statistical shortcuts. The task remains identical to VQA: given an image and a question, generate the most accurate natural-language answer, which is evaluated with the standard consensus-based soft accuracy metric [19].

#### 3) WHAT IT MEASURES (BEYOND ACCURACY)

VQA-CP provides a controlled testbed for studying: (1) a model's sensitivity to distribution shifts in language priors, (2) its ability to perform visual grounding rather than text-only reasoning, and (3) the effectiveness of bias-reduction techniques such as loss reweighting, adversarial regularization, and attention supervision [29], [66], [67]. Because the only change is in train/test pairing statistics, performance drops directly reveal the extent of shortcut learning.

#### 4) REPRESENTATIVE FINDINGS

Early neural models that performed well on standard VQA showed a dramatic performance collapse under VQA-CP. For instance, the top VQA v2 models in 2018 dropped by more than 20 absolute points when evaluated on VQA-CP splits [29]. This sparked a line of research into debiasing strategies: - *Adversarial regularization*: GRAD [68] trains an adversary to detect question-type priors, thereby pushing the base model toward balanced reasoning. - *Attention-based grounding*: HINT [67] supervises visual attention maps using human annotations to encourage grounding. - *Data augmentation and balancing*: RUBi [69] and LMH [66] downweight question-only signals during training. While modern transformer-based architectures such as LXMERT [63] and UNITER [70] achieve better results, large gaps remain, showing that multimodal models continue to struggle with robustness to priors [71].

#### 5) WHAT IT LACKS (AS AN EVALUATION)

VQA-CP focuses on a single type of robustness challenge: shifting question–answer priors. While this is crucial for diagnosing bias, it does not test other sources of distribution shifts, such as novel visual concepts or compositional generalization. As a result, VQA-CP is best interpreted as a diagnostic complement to broader benchmarks, such as GQA out-of-distribution (GQA-OOD) [60] or NLVR2, rather than a standalone measure of reasoning ability.

The structure and evaluation protocol of the VQA-CP benchmark, designed to test robustness against language priors and shortcut learning in Visual Question Answering, are summarized in Table 7.

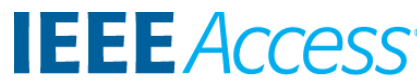

### C. GQA: COMPOSITIONAL VISUAL REASONING WITH PROGRAM SUPERVISION

#### 1) WHAT IT IS

GQA [26] is a large-scale benchmark designed to evaluate compositional visual reasoning by pairing images with questions grounded in scene graphs and accompanied by functional programs. Unlike earlier VQA-style datasets, GQA explicitly structures questions into compositional operations, building on the paradigm first introduced by CLEVR [30] and modular reasoning approaches such as Neural Module Networks [72]. For instance, the question "What color is the sofa on the left?" corresponds to a reasoning chain such as *select(sofa)* $\rightarrow$ *relate(left)* $\rightarrow$ *query(color)*, requiring the model to locate all sofas, identify the one on the left, and then query its color. This structure enables systematic evaluation of whether models can combine concepts through multi-step reasoning rather than relying on shallow correlations.

#### 2) COVERAGE AND DESIGN

GQA leverages densely annotated scene graphs from Visual Genome [73] to generate over *22 million compositional questions* programmatically. Each question is linked to a functional program and precise scene graph regions, supporting a fine-grained diagnostic analysis of visual grounding and reasoning steps. Functional programs also give dataset creators tight control over statistical properties, enabling careful balancing of answer distributions across question groups. This bias mitigation strategy follows lessons from prior work such as VQA-CP [29] and broader shortcut learning research [20], making it difficult for models to succeed by exploiting language priors instead of genuine visual reasoning.

In addition to a balanced standard split, GQA introduces a suite of diagnostic metrics that go beyond simple accuracy to provide a multi-dimensional view of a model's reasoning behavior: - *Accuracy*: Standard measure of correctness on the balanced test set. - *Consistency*: Measures whether a model provides contradictory answers to logically entailed questions. For instance, if a model answers "Yes" to "Is the apple to the left of the plate?", it should also correctly answer "Is the plate to the right of the apple?". - *Validity*: Checks whether answers are well-formed given the question type. - *Plausibility*: Evaluates whether answers are realistic in context, even if incorrect. - *Grounding*: For attention-based models, evaluates whether the model attends to the correct image regions when answering a question, using techniques such as Grad-CAM [74]. - *Distribution*: Compares the model's predicted answer distribution to the true distribution, revealing whether it can handle rare, long-tail answers instead of only frequent ones [75].

By introducing this multi-faceted evaluation framework, GQA's designers effectively acted as psychometricians, creating a benchmark that provides a detailed diagnostic profile of a model's reasoning abilities rather than a simple pass/fail grade.

**TABLE 8.** GQA benchmark at a glance.

| Property | Value |
|---|---|
| Modalities | Image + Text (functional-program-grounded questions) |
| Design basis | Visual Genome scene graphs with entities, attributes, relations |
| Scale | 22M questions over 113k images |
| Tasks | Open-ended question answering (QA) with compositional reasoning chains (filter, relate, count, compare, aggregate) |
| Primary metrics | Accuracy; diagnostics: consistency, validity, plausibility, grounding, distribution |
| Diagnostic focus | Compositionality, relational reasoning, multi-hop inference, bias reduction |
| Robustness variants | OOD splits and rebalanced distributions [60] |

#### 3) WHAT IT MEASURES (BEYOND ACCURACY)

GQA directly probes *compositionality*, *relational reasoning*, and *multi-hop inference* by decomposing questions into sequences of reasoning primitives. Program supervision enables the measurement of whether models solve the right subproblems for the right reasons (e.g., filtering entities before counting). Auxiliary metrics reveal hidden failure modes in which raw accuracy might be concealed, for example, a model could achieve high accuracy while being logically inconsistent or relying on dataset shortcuts.

#### 4) REPRESENTATIVE FINDINGS

(1) Models that perform well on VQA often struggle with GQA compositional queries, revealing a persistent gap between surface pattern recognition and true reasoning. (2) Transformer-based multimodal encoders have improved aggregate accuracy, but diagnostics such as consistency, grounding, and plausibility frequently lag behind headline numbers, thereby exposing brittle reasoning chains. (3) Under deliberate distribution shifts—such as altered attribute or relation frequency performance degrades sharply [60], underscoring the difficulty of systematic generalization and the prevalence of spurious correlations [54], [71].

#### 5) WHAT IT LACKS (AS AN EVALUATION)

While GQA's structured design and diagnostic tools of GQA represent a significant advancement, its templated question generation can still introduce stylistic regularities that models may partially exploit. Additionally, noise in the underlying scene graphs can propagate errors, and success of GQA does not guarantee robustness to real-world images or open-world knowledge integration.

GQA's compositional annotations and evaluation settings are outlined in Table 8.

### D. CLEVR AND CLEVR-CoGenT: SYNTHETIC COMPOSITIONAL REASONING AND GENERALIZATION

#### 1) WHAT IT IS

CLEVR [30] is a synthetic benchmark explicitly designed to diagnose *compositional* visual reasoning. Images are procedurally generated along with ground-truth scene graphs, and each question is paired with a functional program (e.g., *filter(color=red)* $\rightarrow$ *relate(left)* $\rightarrow$ *count*) that specifies the precise sequence of reasoning operations required. The *CoGenT* (Compositional Generalization Test) split [30] extends CLEVR to test whether models can *systematically generalize* to unseen combinations of attributes, using disjoint color–shape pairings across train/test conditions (A/B) while keeping other factors controlled.

#### 2) COVERAGE AND DESIGN

CLEVR provides millions of question–answer pairs rendered from a compact set of primitives (objects: *shape*, *color*, *size*, *material*; relations: *left of*, *behind*, *closest*, etc.). Each question is generated from templates and compiled into a functional program composed of reasoning primitives such as *filter*, *relate*, *count*, *compare_attribute*, and *compare_integer* [30]. This design minimizes linguistic and annotation ambiguity (compared to natural-image VQA) and enables fine-grained, category-specific diagnostics (e.g., *counting*, *comparing numbers*, *same–different*, and*existence*). The CoGenT split [30] imposes disjoint color–shape pairings between the train and test (Condition A $\leftrightarrow$ B; spheres remain unconstrained), creating a controlled out-of-distribution setting specifically targeted at *systematic* generalization rather than mere interpolation.

#### 3) WHAT IT MEASURES (BEYOND ACCURACY)

CLEVR probes whether models can execute *multi-step, compositional* reasoning chains grounded in vision: filtering by attributes, reasoning over spatial relations, aggregating sets, and performing comparisons. Because questions are program-supervised, evaluation can analyze where a failure originates (for example, incorrect attribute filtering vs. relational reasoning). The CoGenT split measures whether a model's learned representations factorize attributes (e.g., *color* vs. *shape*) such that it can recombine them at the test time that is, *systematic compositional generalization* [30]. Together, CLEVR and CoGenT move evaluation beyond surface answer accuracy to test whether the models solve the *right subproblems for the right reasons* [72].

#### 4) REPRESENTATIVE FINDINGS

(1) Early Convolutional Neural Network (CNN) and Long Short-Term Memory (LSTM) baselines struggled, while architectures explicitly designed for relational and compositional reasoning, for example, Relation Networks (RN) [76], Neural Module Networks and successors [72], [77], and program-aware neuro-symbolic approaches such as NS-VQA [78]—achieved large gains on CLEVR. (2) Conditioning mechanisms that modulate visual features with language, such as FiLM [79], and recurrent multi-step controllers such as the MAC network [80], further improved data efficiency and reasoning performance on compositional categories. (3) Nevertheless, many models that reached near-saturated accuracy on standard CLEVR exhibited substantial decreases on CoGenT A$\rightarrow$B generalization, revealing limited attribute factorization and heavy reliance on training co-occurrence statistics [30], [79], [80].

**TABLE 9. CLEVR and CLEVR-CoGenT at a glance.**

| Property | Value |
| --- | --- |
| Modalities | Synthetic Image + Text (program-grounded questions) |
| Design basis | Procedural scene graphs; functional programs of reasoning primitives [30] |
| Diagnostic tasks | Filter, relate, count, compare attributes/integers; existence; same–different |
| CoGenT shift | Disjoint color–shape pairings across train/test (Conditions A/B); spheres unconstrained [30] |
| Primary metrics | Category-wise accuracy (e.g., count, compare, exist) with per-type diagnostics |
| Notable model families | RN [76]; FiLM [79]; MAC [80]; NS-VQA [78] |
| Core limitation | Synthetic domain and templated language; limited ecological validity |

(4) Follow-ups showed that stronger *program supervision*, disentangling objectives, and targeted data augmentation can mitigate (but not eliminate) failures in CoGenT, underscoring the difficulty of *systematic* generalization in vision–language reasoning [78], [80].

#### 5) WHAT IT LACKS (AS AN EVALUATION)

Despite its diagnostic clarity, CLEVR is synthetic: images are simple 3D renderings and language is templated, which limits ecological validity and can yield stylistic regularities that models overfit. High CLEVR scores do not guarantee robustness on natural images, noisy optical character recognition (OCR), or open-world knowledge. Moreover, multiple-choice variants are easier than free-form generation, and even in CoGenT, the distribution shifts are controlled along a single axis (attribute pairing), leaving broader real-world shifts (lighting, clutter, camera, and domain semantics). As such, CLEVR/CoGenT is best viewed as a *unit test* for compositional reasoning combined with natural-image benchmarks.

Diagnostic design choices in CLEVR (compositional templates and functional programs) appear in Table 9.

### E. NLVR2: NATURAL-IMAGE COMPOSITIONAL REASONING WITH TRUTH-CONDITIONAL JUDGMENTS

#### 1) WHAT IT IS

The Natural Language for Visual Reasoning (NLVR) benchmark [31] introduced a new evaluation paradigm in which a system must determine whether a natural-language statement is *true* with respect to a set of images. NLVR uses synthetic images with controlled visual elements, enabling precise testing of logical operators, counting, and spatial reasoning. While it successfully highlighted the challenges in grounding language to visual structure, its synthetic nature limited ecological validity and encouraged models to exploit stylistic artifacts rather than performing robust reasoning.

NLVR2 [27] was proposed to address these limitations by moving from synthetic to *real-world photographs*, significantly increasing the visual complexity and diversity. Each example pairs a sentence with two photographs, which requires joint reasoning across them under explicit truth conditions. This naturalistic setting makes NLVR2 a more challenging and realistic benchmark for compositional visual reasoning, bridging the gap between synthetic tasks such as CLEVR and natural-image benchmarks such as GQA.

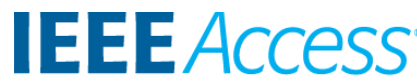

#### 2) COVERAGE AND DESIGN

NLVR2 contains approximately 107k sentence–image-pair examples. The dataset was built in three stages: (1) collecting visually rich and diverse photo sets, (2) eliciting crowd-authored statements that were *true* for some image pairs and *false* for others, and (3) validating the statements across multiple pairings [27]. Each sentence appears in contexts where it has both labels, a design choice that substantially reduces pure language priors and forces models to use visual evidence rather than textual bias. The evaluation is based on binary accuracy, supplemented by a *consistency* metric that aggregates the correctness across all occurrences of the same sentence. Subsequent analyses introduced additional subsets to monitor potential visual biases and assess their robustness [81].

#### 3) WHAT IT MEASURES (BEYOND ACCURACY)

NLVR2 probes a model's ability to perform *truth-conditional* reasoning with natural images across four key dimensions: (1) *set and quantity reasoning*, such as identifying numerical relationships (e.g., ‘‘twice as many dogs as cats’’); (2) *logical compositionality*, including conjunctions, negations, and quantifiers; (3) *cross-image relational reasoning*, where two panels must be compared to determine truth conditions; and (4) *visual grounding*, since each sentence is presented in both true and false contexts to discourage language priors. By systematically pairing sentences with multiple labeled image pairs, NLVR2 minimizes shortcut learning and emphasizes robust, integrated vision–language understanding [27].

#### 4) REPRESENTATIVE FINDINGS

(1) Initial VQA-style baselines and program-induction models achieved only modest performances, demonstrating the difficulty of naturalistic, set-theoretic reasoning [27]. (2) Vision–language pretraining leads to large gains: LXMERT improved the performance from ∼54% to ∼76% via cross-modal transformers [63], while UNITER further advanced the state of the art with fine-grained word-region alignment and optimal transport objectives [70]. OSCAR's object-tag pretraining and VinVL's enhanced detectors continued to increase performance [82], [83]. (3) Despite improvements, consistency remains lower than headline accuracy, revealing instability across repeated sentences and continued difficulty with complex compositional logic [27].

#### 5) WHAT IT LACKS (AS AN EVALUATION)

NLVR2's binary setup simplifies scoring but cannot capture nuanced partial correctness or generate free-form explanations. Although its design reduces the number of language priors, some visual biases and artifacts remain [81]. Moreover, while NLVR2 focuses on compositional perception and logical reasoning, it does not test knowledge-grounded inference (cf. OK-VQA) or justification using natural language (cf. VCR).

**TABLE 10. NLVR2 benchmark at a glance.**

| Property | Value |
|---|---|
| Modalities | Natural images (pairs) + text (truth-conditional statements) |
| Task | Decide if the sentence is true about the *pair* of photos |
| Scale | ∼107k examples; each sentence occurs with both true and false labels [27] |
| Metrics | Accuracy; sentence-level *consistency* across contexts [27] |
| Diagnostic focus | Logical operators, counting, set relations, cross-image comparisons, visual grounding |
| Notable results | LXMERT [63], UNITER [70], OSCAR [82], VinVL [83] |
| Known limitations | Binary outputs, residual visual bias, limited justification or external knowledge probing |

Core evaluation protocol and splits for NLVR2 are summarized in Table 10.

### F. WINOGROUND: DIAGNOSING MULTIMODAL BINDING FAILURES

#### 1) WHAT IT IS

While benchmarks such as CLEVR, GQA, and NLVR2 evaluate compositional reasoning and relational understanding, they do not fully capture a model's ability to *bind* language to correct visual referents. Winoground [84] was introduced to explicitly diagnose this issue by testing whether models can distinguish subtle differences in language–vision pairings. Inspired by the Winograd Schema Challenge [85], Winoground extended the idea of minimal pairs to multimodal inputs, providing a direct test of whether a model truly integrates linguistic structures with visual grounding rather than performing shallow pattern matching.

#### 2) COVERAGE AND DESIGN

The benchmark consisted of 400 challenging examples, each composed of two images and two captions. For each example, one caption correctly describes the first image and the other correctly describes the second image, with the remaining two pairings being incorrect. This creates four possible image–caption combinations, only two of which are correct. For a model to succeed, it must simultaneously: (1) assign a higher score to the correct caption for each image, and (2) correctly match each caption to its intended image. This setup forces models to attend to precise relational cues, such as word order or role assignment, rather than relying on unigram overlap or object co-occurrence [84].

#### 3) WHAT IT MEASURES (BEYOND ACCURACY)

Winoground is designed to evaluate *multimodal binding*, that is, whether a system can correctly link linguistic elements (nouns, relations, and attributes) to the corresponding visual regions. It emphasizes: (1) *Relational reasoning*, such as distinguishing between ‘‘the dog chasing the cat’’ vs. ‘‘the cat chasing the dog’’; (2) *Compositional generalization*, testing whether models understand novel combinations of familiar words and objects; and (3) *Cross-modal grounding*, ensuring that success cannot be achieved by language-only or vision-only priors. This directly operationalizes the ‘‘binding problem’’ described in cognitive science.

**TABLE 11. Winoground benchmark at a glance.**

| Property | Value |
|---|---|
| Modalities | Natural images + text (pairs of images and captions) |
| Task | Correctly match each caption to its corresponding image across four possible pairings |
| Scale | 400 challenging examples with minimal linguistic and visual differences [84] |
| Metrics | Accuracy on strict pair matching; caption-level and image-level accuracy |
| Diagnostic focus | Multimodal binding, relational grounding, role assignment, compositional generalization |
| Known limitations | Small dataset size, limited domains, no partial-credit scoring |

#### 4) REPRESENTATIVE FINDINGS

State-of-the-art models at the time of Winoground's release, including Contrastive Language–Image Pretraining (CLIP) [86], ALIGN [87], and UNITER [70], performed at or near random-chance levels, with accuracy below 10% on the strict matching metric [84]. Subsequently, large multimodal models, such as Flamingo [88] and GPT-4V [89], have shown measurable but still limited improvements, highlighting that even cutting-edge architectures struggle to achieve robust multimodal alignment. Analysis revealed characteristic failure modes: models often identify the correct objects but swap their roles or relations, echoing the classic Winograd schema challenges in language-only reasoning [85].

#### 5) WHAT IT LACKS (AS AN EVALUATION)

While Winoground is highly diagnostic, its small size (400 examples) limits statistical power and generalization. It serves primarily as a "unit test" for multimodal grounding rather than as comprehensive benchmark. Additionally, its binary matching format provides no insight into partial correctness or reasoning process behind a model's choice. Consequently, Winoground is best used alongside larger benchmarks such as NLVR2 or GQA to contextualize its fine-grained diagnostic signals.

### G. OK-VQA AND A-OKVQA: KNOWLEDGE-GROUNDED VISUAL QUESTION ANSWERING

#### 1) WHAT IT IS

While traditional VQA benchmarks [19], [56] focus on perception and reasoning within an image, many real-world questions require *external knowledge* that is not explicitly visible. For example, answering "Why are people wearing helmets?" requires knowledge of biking safety, and not just the recognition of helmets or people. The Outside Knowledge Visual Question Answering (OK-VQA) benchmark [58] was created to explicitly test this capability by requiring models to integrate visual understanding with world knowledge sources such as commonsense, facts, and cultural context. Its successor, Augmented OK-VQA(A-OKVQA) [59], expands both scale and task diversity, providing a richer and more challenging benchmark for the emerging era of multimodal large language models (MLLMs).

#### 2) COVERAGE AND DESIGN

OK-VQA consists of ∼14,000 images and ∼14,000 open-ended questions sampled from MS-COCO, with a deliberate design to ensure that solving each question requires external knowledge beyond the image. The dataset spans 11 high-level knowledge categories: science, sports, geography, culture, and history. Each question was paired with multiple human-provided answers to allow consensus-based evaluation [58]. Building on this foundation, A-OKVQA introduces two major innovations: (1) a *much larger scale* with over 25,000 questions and more diverse answer types, and (2) a shift from single-task evaluation to a *multitask framework*, supporting both classification-style multiple choice and open-ended generation. This design allows the same benchmark to be used for the training and evaluation of modern foundation models.

#### 3) WHAT IT MEASURES (BEYOND ACCURACY)

OK-VQA and A-OKVQA probe a model's ability to: (1) extract relevant visual cues from an image, (2) query or retrieve relevant external knowledge (e.g., through pretrained LLMs or symbolic knowledge bases), and (3) integrate perception with knowledge to generate a coherent answer. These steps mirror how humans combine what they *see* with what they *know*. Therefore the benchmark is critical for diagnosing whether a model's reasoning is grounded only in surface-level perception or truly augmented by real-world understanding [59]. By including multiple knowledge domains, it also exposes weaknesses in the generalization across topics.

#### 4) REPRESENTATIVE FINDINGS

Initial studies found that early VQA architectures such as UpDn and BAN [62], [90] performed poorly on OK-VQA, often near random chance in many categories [58]. Subsequent models that integrate explicit retrieval or pretrained language models, such as KRISP [91] and MMBERT [92], significantly improved the results by combining visual features with structured knowledge bases or Wikipedia-scale corpora. A-OKVQA has become a standard evaluation for modern multimodal transformers such as Flamingo [88] and GPT-4V [89], which achieve far higher scores than VQA-specific baselines but still struggle with rare or niche knowledge categories. Analyses reveal persistent failure modes: models often hallucinate plausible but incorrect answers, or fail to integrate retrieved information into reasoning chains [59].

#### 5) WHAT IT LACKS (AS AN EVALUATION)

Although OK-VQA and A-OKVQA are vital for knowledge-grounded reasoning, they have limitations. OK-VQA's small size of OK-VQA restricts its statistical robustness and makes it vulnerable to overfitting. A-OKVQA addresses this to some extent with scale and multitask design, but both benchmarks remain biased toward MS-COCO imagery and English-language cultural assumptions. Additionally, their open-ended nature complicates scoring, as multiple valid answers may exist, but be penalized if not matched to the reference set [59]. Thus, although these benchmarks are essential diagnostic tools, they are best used alongside perception-focused datasets such as GQA and grounding-focused tests such as Winoground.

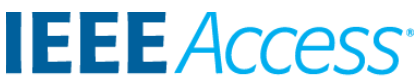

**TABLE 12. OK-VQA and A-OKVQA benchmarks at a glance.**

| Property | Value |
|---|---|
| Modalities | Natural images + text (open-ended questions and answers) |
| Design basis | Questions require external knowledge beyond the image; A-OKVQA adds multitask evaluation |
| Scale | OK-VQA: 14k Qs over 14k images; A-OKVQA: 25k+ Qs with expanded domains |
| Metrics | Consensus-based soft accuracy; multiple-choice and generative evaluation modes |
| Diagnostic focus | Knowledge grounding, retrieval integration, cross-domain reasoning |
| Knowledge domains | 11+ categories including science, culture, geography, history, sports |
| Known limitations | Small size (OK-VQA), cultural bias, ambiguous answer evaluation |

The characteristics and evaluation protocols of the OK-VQA and A-OKVQA benchmarks, which extend traditional VQA by emphasizing external knowledge requirements, are summarized in Table 12.

### *H. VISUAL COMMONSENSE REASONING (VCR): DEMANDING JUSTIFICATION*

#### 1) WHAT IT IS

Building on benchmarks that test perception (VQA), compositional reasoning (GQA, NLVR2), grounding (Winoground), knowledge integration (OK-VQA/A-OKVQA), and the Visual Commonsense Reasoning (VCR) benchmark [28] pushes multimodal models toward higher-level cognition. True understanding requires inferring the unstated context of a scene—the intentions, emotions, and causal relationships between entities [93], [94]. The VCR was introduced to explicitly test this ability, moving beyond object recognition and spatial reasoning to evaluate whether a model can combine visual cues with external commonsense knowledge.

#### 2) COVERAGE AND DESIGN

The VCR dataset was derived from over 110,000 diverse movie scenes, annotated with complex questions targeting higher-level understanding. This design introduces three key innovations:

- *Two-stage reasoning task:* A model is first presented with an image and a natural-language question and must select the correct answer from four choices ($Q \rightarrow A$ task). Crucially, it must select the correct rationale from another set of four options that justifies why the chosen answer is correct ($QA \rightarrow R$ task). This staged evaluation process mimics a two-step exam: first answering and explaining reasoning.
- *Holistic evaluation metric:* For full credit, a model must answer both questions correctly ($Q \rightarrow AR$). This penalizes lucky guesses: random chance for $Q \rightarrow A$ is 25%, but for $Q \rightarrow AR$ it drops to 6.25%, making shortcut exploitation extremely unlikely [28].
- *Adversarial Matching:* To prevent annotation artifacts, the creators developed an adversarial data collection procedure in which incorrect distractor options were automatically chosen to be highly plausible yet incorrect [28]. This significantly reduces exploitable biases and forces models to engage in deeper reasoning, a principle later adopted in other datasets such as SWAG [95] and Social IQa [96].

The VCR's structure explicitly links visual perception and textual justification, making it one of the first benchmarks to operationalize the notion of 'showing your work' in multimodal reasoning.

#### 3) WHAT IT MEASURES (BEYOND ACCURACY)

VCR directly tests a model's ability across four dimensions: (1) *recognition of entities and interactions*, including people, objects, and their relationships; (2) *causal inference*, which requires the model to reason about unstated relationships and intentions; (3) *integration of commonsense and external knowledge* beyond what is visually depicted; and (4) *justification of decisions* through coherent natural-language rationales. Approximately 38% of VCR questions are explicitly explanatory ("Why" or "How") [28], which makes it uniquely positioned to target higher-level cognitive reasoning compared to traditional visual QA tasks. This aligns VCR with cognitive science-inspired approaches to model evaluation, bridging perception and cognition [97].

#### 4) REPRESENTATIVE FINDINGS

The benchmark exposed a stark performance gap: humans achieve over 90% accuracy on the full $Q \rightarrow AR$ metric, whereas the accuracy of baseline R2C model proposed by the dataset creators reached only approximately 65% [28]. Subsequent transformer-based approaches, such as ViLBERT [64] and LXMERT [63], improved the raw accuracy, but often failed to provide logically consistent rationales [71]. Analysis of model outputs revealed that despite architectural advances, many systems continue to rely on surface-level correlations rather than genuine causal inference or commonsense integration [54]. These findings cemented VCR's role of VCR as a diagnostic benchmark for probing the gap between visual perception and reasoning.

#### 5) WHAT IT LACKS (AS AN EVALUATION)

Although VCR's two-stage design of the VCR represents a major step forward, reliance on multiple-choice options limits its ecological validity. Generating free-form rationales remains far more challenging and realistic than selecting predefined rationales [98]. Additionally, because VCR is built primarily from movie scenes, it may underrepresent other domains such as scientific imagery or real-world photographs. Consequently, while VCR is a cornerstone benchmark for multimodal commonsense reasoning, it is best used in combination with datasets, which target diverse contexts. By bridging perception, grounding, and knowledge integration, VCR also serves as a natural transition point

**TABLE 13.** **VCR benchmark at a glance.**

| **Property** | **Value** |
|---|---|
| Modalities | Image + Text (questions, answers, rationales) |
| Scale | 110k movie scenes, ∼290k questions with rationales |
| Tasks | Two-stage reasoning: $Q \rightarrow A$ (answering), $QA \rightarrow R$ (rationale selection) |
| Primary metric | $Q \rightarrow AR$ joint accuracy (both stages correct) |
| Diagnostic focus | Commonsense inference, causal reasoning, justification |
| Bias mitigation | Adversarial Matching for hard negative distractors |

to Level 3 benchmarks, which focus on complex and open-ended forms of reasoning.

We detail VCR's question types, rationales, and evaluation metrics in Table 13.

### *I. DIAGNOSTIC TESTING FOR SYSTEMIC FLAWS*

The evolution of Level II benchmarks not only expanded the scope of multimodal evaluation but also catalyzed a deeper understanding of *why* models fail. The creation of diagnostic benchmarks, such as GQA, VQA-CP, Winoground, and VCR, coincided with a research movement to formally define and categorize the systemic flaws they were designed for exposure. These flaws represent fundamental failure modes in deep learning systems, revealing the gap between high performance on independent and identically distributed test sets and robust, and generalizable intelligence [20], [54]. By explicitly engineering challenging conditions, diagnostic benchmarks serve as stress tests that illuminate the boundaries of current multimodal reasoning.

#### 1) SHORTCUT LEARNING AND SPURIOUS CORRELATIONS

Shortcut learning occurs when a model relies on simple, spurious correlations in the training data instead of learning the intended, robust features of a task [99]. These shortcuts are often effective for minimizing loss in the training distribution but fail catastrophically when that distribution shifts, as it inevitably occurs in real-world settings [54], [100].

Classic examples from computer vision vividly illustrate this concept:

- *Context as a Shortcut:* A model trained to classify cows may learn to associate the presence of green grass with the ''cow'' label. When presented with an image of a cow on a beach, it fails because the shortcut feature (grass) is absent [20].
- *Dataset Artifacts as Shortcuts:* In a famous early example, a model trained to distinguish U.S. from Russian tanks learned to classify them based on the time of day the photos were taken, a feature completely uncorrelated with the tanks themselves but systematically tied to the data collection process. Similarly, in medical imaging, models trained to detect pneumonia have been found to rely on metal tokens or chest tubes present in X-rays from certain hospitals rather than the actual pathology [54], [55].

In multimodal reasoning, shortcut learning often manifests as a reliance on language priors rather than true visual grounding. For example, early VQA models answered most yes/no questions with ''yes'' because it maximized the accuracy of the training set [65]. To diagnose this behavior, *VQA-CP* was introduced, deliberately *flipping* the distribution of question–answer pairs between the training and test splits [29]. If 80% of the training questions about bananas had the answer ''yellow,'' the test split would invert this ratio to 20%, forcing models to attend to the actual image rather than memorizing statistical co-occurrences. This explicit distribution shift revealed dramatic performance collapses in otherwise high-performing models and spurred new research into debiasing techniques such as RUBi [69] and HINT [67]. Similarly, *GQA-OOD* separates question–answer pairs into ''head'' (frequent) and ''tail'' (rare) distributions, with performance on rare concepts serving as a better measure of true reasoning ability [60].

These benchmarks have shown that shortcut learning is not merely a nuisance but also a fundamental barrier to generalization. Without diagnostic evaluation, models may appear to ''understand'' when in reality they are exploiting the dataset statistics.

#### 2) THE BINDING PROBLEM: FAILURES IN COMPOSITIONALITY

Compositionality is a cornerstone of human intelligence: the ability to understand the meaning of a whole as a function of its parts and the rules for combining them [101]. For example, a model that understands ''red,'' ''cube,'' ''blue,'' and ''sphere'' should be able to correctly interpret the novel phrase ''a red cube next to a blue sphere'' [102].

However, many vision–language models fail at this integration step, exhibiting what is known as the *binding problem*. They may correctly detect individual objects and attributes but fail to bind them into coherent relational structures. This leads to the following characteristic failure modes.

- *Attribute Swapping:* A model showing an image of ''a red cube and a blue sphere'' might incorrectly validate the caption ''a blue cube and a red sphere'' because it identifies the components but not their correct assignments [103].
- *Relation Errors:* Given the prompt ''a man riding a horse,'' a model might generate or validate an image of a man standing next to a horse, failing to capture the specified action or relationship [104].
- *Negation Misinterpretation:* Many models exhibit *affirmation bias* and incorrectly interpret negated statements. Benchmarks such as NegBench [105] and NegVQA [106] reveal substantial drops in accuracy on negated prompts such as ''no dog in the image.''
- *Counting Errors:* Numerical reasoning remains fragile: models often detect object presence but fail at precise quantification, e.g., validating ''three apples'' when there are only two [107].

Diagnostic benchmarks such as *CLEVR-CoGenT* [30] and *Winoground* [84] directly target compositionality and binding. CLEVR-CoGenT enforces disjoint attribute combinations between the training and test sets, testing whether models can recombine familiar parts in novel ways. Winoground uses natural images and paired captions to probe fine-grained alignment, requiring models to distinguish between relational role reversals such as "the dog chasing the cat" vs. "the cat chasing the dog." Success on these tasks requires systematic generalization beyond more pattern recognition.

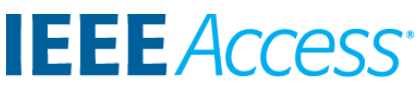

#### 3) ADVERSARIAL FRAGILITY

A third major weakness exposed by diagnostic testing is *adversarial fragility*: the susceptibility of models to small perturbations that humans easily ignore but which causes dramatic model failures. While early works focused on imperceptible noise, a more revealing modern approach is *adversarial data collection*, where humans are actively involved in creating failure cases.

The *Adversarial VQA (AdVQA)* benchmark exemplifies this methodology [53]. Human annotators interact with a state-of-the-art VQA model and deliberately craft questions that the model answers incorrectly [108]. For instance, a model may correctly answer "How many cats are in the image?" but fail on the more nuanced question "How many cartoon drawings are present on the cat's tie?" [53].

This approach has several advantages:

- *Moving Target:* As the models improve, new adversarial data can be collected, ensuring that the benchmark remains challenging and avoids saturation.
- *Uncovering Unknown Unknowns:* Humans excel at generating unexpected failure cases, such as rare counting tasks, subtle text-reading errors, or complex commonsense inferences that automated perturbation methods would miss.

The adversarial benchmarks demonstrate that even when a model achieves near-human performance on static datasets, its understanding remains brittle and fragile. These findings underscore the need for continuous stress testing to prevent overfitting in narrow evaluation regimes.

#### 4) KNOWLEDGE GAPS AND EXTERNAL REASONING

A final limitation of recent benchmarks is the lack of integration between perception and external world knowledge. While models may excel at identifying objects and spatial relationships, they often fail when answering questions that require commonsense or encyclopedic information. Benchmarks such as *OK-VQA* [58] and its successor *A-OKVQA* [59] explicitly test this integration by designing questions whose answers cannot be derived from the image alone. For example, identifying a picture of people in helmets may be easy, but explaining "Why are they wearing helmets?" requires contextual knowledge about biking safety or construction work.

Performance gaps on these benchmarks highlight that visual reasoning remains incomplete without mechanisms for knowledge retrieval and integration. Even state-of-the-art multimodal LLMs such as GPT-4V show measurable improvements but still struggle with rare or specialized knowledge domains [89].

### J. SYNTHESIS OF LEVEL II: THE ALIGNMENT ERA

The Alignment Era (VQA [19], GQA [26], VCR [28]) represented a necessary adversarial response to the brittleness of the previous generation. As synthesized in Table 14, the defining characteristic of this era was a dramatic surge in the *Robustness* rating. By deliberately engineering distribution shifts (VQA-CP) and compositional stress tests (CLEVR), researchers prioritized diagnostic rigor over raw performance. This sophistication, however, introduced significant trade-offs: *Reliability* scores dipped as metrics struggled to handle open-ended synonyms, and *Hygiene* ratings plummeted as the field discovered that models were massively overfitting to "language priors"—statistical correlations in the training data that allowed them to bypass visual reasoning entirely [65]. This failure of static datasets necessitated the shift to *Level III*, where expert-level tasks and open-ended generation attempt to force genuine reasoning over pattern matching.

## V. LEVEL III: EXPERT-LEVEL MULTIMODAL INTEGRATION (THE CURRENT FRONTIER)

The advent of Multimodal Large Language Models (MLLMs)—unified architectures capable of processing and reasoning over interleaved streams of text, images, video, and audio—has inaugurated the current era of AI evaluation, from approximately the 2020s to the present [109]. These powerful models, which build upon the foundations of large language models, demand a new class of "holistic exams" that go beyond testing isolated skills. The current frontier of benchmarking focuses on assessing the ability of MLLMs to synthesize information across modalities, apply deep domain knowledge, and articulate complex reasoning processes. This has led to a landscape of evaluation tools that can be broadly categorized into two complementary types: *broad-spectrum competency exams* designed to assess a wide range of abilities, and *deep-dive diagnostic tests* that probe a single, complex capability in extreme detail [2], [2].

### A. THE RISE OF MLLMs AND THE NEED FOR HOLISTIC EXAMS

Models such as OpenAI's GPT-4V, Google's Gemini, and open-source alternatives have demonstrated remarkable capabilities, excelling in many existing multimodal benchmarks [21]. However, their very generality presents a challenge for evaluation. A single task, such as object detection or VQA, is insufficient for capturing the breadth of their skills. Consequently, the research community has moved toward creating comprehensive benchmarks that function as "final exams," testing a wide array of competencies required for expert-level performance in real-world scenarios [2]. This includes not only broad-spectrum exams but also deep-dive diagnostics that probe fundamental skills, such as fine-grained pattern recognition [110] or basic object counting [111], where MLLMs surprisingly still struggle. These exams are characterized by their diversity in tasks, modalities, and complexity of reasoning required.

**TABLE 14.** **Framework synthesis: Level II (Alignment & Reasoning).**

| Dimension | Rating | Justification and Evidence |
|---|---|---|
| *Skill* | ⋆⋆⋆ | *Moderate.* Expanded scope to compositionality and commonsense. GQA introduced 22M compositional questions [26], and VCR required rationale generation [28], though domains remained largely closed (Visual Genome). |
| *Reliability* | ⋆⋆⋆ | *Mixed.* Introduced "Soft Accuracy" to handle annotator disagreement [19], but exact-match metrics penalize valid synonyms, introducing scoring noise absent in Level I classification. |
| *Robustness* | ⋆⋆⋆⋆ | *High.* Explicitly designed as stress tests. VQA-CP flipped answer distributions to penalize shortcut learning, causing State-of-the-Art (SOTA) model accuracy to drop by over 20% [29]. |
| *Hygiene* | ⋆⋆ | *Low (Language Priors).* Severe intrinsic contamination. Analysis revealed models could achieve ∼60% accuracy on VQA v1 without looking at the image, relying solely on question statistics [20], [65]. |
| *Cost* | ⋆⋆⋆ | *Moderate.* The shift to Vision-Language Pretraining (e.g., ViLBERT [64], LXMERT [63]) increased compute requirements relative to CNNs, but remained accessible compared to Level III LLMs. |
| *Fairness* | ⋆⋆⋆ | *Improving.* Explicit bias-reduction metrics introduced in VQA-CP. Introduction of the *VizWiz* dataset [57] marked the first major effort to evaluate AI on images taken by visually impaired users. |

### B. VIDEO-MME: FULL-SPECTRUM EVALUATION FOR TEMPORAL, AUDIO-AWARE VIDEO REASONING

As models become proficient at static imagery, the evaluation frontier shifts to the temporal dimension. We begin with *Video-MME*, which represents the push for long-horizon reasoning and multi-modal fusion (audio/subtitles), addressing the key challenge of maintaining coherence over time.

#### 1) WHAT IT IS

While most multimodal evaluations concentrate on static images, *Video-MME* targets sequential video understanding with explicit control over *duration*, *domain diversity*, and *modalities*. It comprises *900* carefully curated videos (*256* hours) with *2700* multiple-choice QA items (three per video), spanning *six* primary domains and *30* subfields. Durations range from *11 seconds to 1 hour*, and inputs include video frames plus *subtitles* and *audio* to probe fusion beyond vision-only cues [35], [112].

#### 2) COVERAGE AND DESIGN

Videos are stratified by *duration* (short/medium/long) and sampled across *six domains* with *30* finer-grained subfields to balance topic and temporal diversity. For each video, expert annotators author *three* multiple-choice questions with *plausible distractors* that require temporal grounding rather than single-frame cues. Items are double-checked to ensure that answering correctly typically benefits from *multi-frame evidence* and, where applicable, *subtitle* or *audio* content. The evaluation pipeline explicitly supports *modality toggles* (frames only; frames + subtitles; frames + audio; frames + subtitles + audio), enabling controlled ablations and per-split reporting by *domain* and *duration* [35], [112].

#### 3) WHAT IT MEASURES

Video-MME stresses (i) *temporal compositionality* (events across minutes to an hour), (ii) *long-context memory* (S/M/L split by duration), and (iii) *audio-/subtitle-grounded reasoning*. Items are manually authored and reviewed by expert annotators, reducing label noise and template leakage. The primary metric is 4-option multiple-choice *accuracy*, reported overall and by duration split, with ablations that toggle the availability of subtitles and audio [35].

#### 4) REPRESENTATIVE FINDINGS

Closed-source frontier models lead but still degrade with longer clips. For example, *Gemini 1.5 Pro* achieves *75.7%* overall without subtitles (*81.6%* with subtitles), whereas a strong open-source video model (*LLaVA-NeXT-Video*) attains *52.5%* (*56.0%* with subtitles). Subtitles and audio provide consistent gains, particularly for longer videos [35]. These results underscore: (a) persistent temporal-reasoning gaps, and (b) the value of multimodal fusion beyond frames alone.

#### 5) WHAT IT LACKS (AS AN EVALUATION)

- *Single headline metric versus multi-metric/process reporting.* Accuracy alone hides trade-offs in calibration, robustness, bias, efficiency, and rationale quality; contrast with HELM-style multi-metric reporting and transparency [1] and with process-level scoring in rationale benchmarks such as VCR-Bench (chain-of-thought (CoT) precision/recall) [113].
- *Limited robustness protocols.* Video-MME does not include systematic controls for instruction sensitivity (format/order effects), choice-order bias, or hallucination diagnostics; compare with *CircularEval* in MMBench and category-wise stressors/human-arena diagnostics in large vision–language model (LVLM)-eHub [17], [114], [115].
- *Static, non-adversarial setting.* Items are curated once; there is no living/adversarial refresh to keep pace with model improvements or surface unknown-unknowns.

**TABLE 15. Video-MME at a glance.**

| Stat / Property | Value |
|---|---|
| Videos / Hours / QA | 900 / 256 / 2700 (3 per video) |
| Domains / Subfields | 6 / 30 |
| Duration range / Splits | 11s–1h / S, M, L |
| Modalities | Frames + Subtitles + Audio |
| Primary metric | MCQ Accuracy (4-option; 0-shot) |
| Ablations | With/without subtitles; with/without audio |

**TABLE 16. MathVista at a glance.**

| Property | Value |
|---|---|
| Modalities | Image + Text (charts, diagrams, tables, scanned pages) |
| Scale | Consolidates 28 datasets; adds IQTest, FunctionQA, PaperQA |
| Tasks | Multiple-choice and open-ended QA with normalized answer extraction |
| Primary metric | Accuracy (Normalized Numeric/String Exact Match) [32] |
| Diagnostic focus | Visual parsing + quantitative reasoning (functions, geometry, data interpretation) |

Adversarial/living paradigms (Dynabench, ANLI, AdVQA and RealTimeQA) show sustained difficulty and reveal new failure modes over time [13], [39], [53], [116].

- *No interaction or embodiment.* The benchmark evaluates passive perception/QA, not goal-directed behavior or planning in environments; embodied evaluations (ALFRED; Habitat Challenge) capture these dimensions and expose different errors [35], [42], [117], [118], [119], [120].
- *Contamination/memorization risk.* As with any static, public benchmark sourced from the open web, there is potential exposure via pretraining corpora, and prior work shows that models can memorize evaluation items without audits/hidden splits [37].

The Video-MME benchmark, which evaluates temporal reasoning and multimodal understanding in video contexts, is summarized in Table 15.

### C. MATHVISTA: VISUAL MATHEMATICAL REASONING ACROSS CHARTS, DIAGRAMS, AND TEXT

While generalist models often succeed at natural scene understanding, they frequently struggle with precise quantitative reasoning. *MathVista* is included here to isolate this deficit, focusing on formal logic, geometric parsing, and algorithmic reasoning within visual contexts.

#### 1) WHAT IT IS

*MathVista* is a large, unified benchmark for *visual mathematical reasoning* that merges *28* prior vision–language math datasets and introduces three new subsets (*IQTest*, *FunctionQA*, *PaperQA*) into a single evaluation suite [32]. It targets problems that interleave *images* (e.g., charts, diagrams, geometric figures, tables and scanned pages) with *text*, emphasizing quantitative reasoning beyond plain VQA [121].

#### 2) COVERAGE AND DESIGN

MathVista spans heterogeneous visual sources (plots, scientific figures, worksheets, exam snippets) and question types ranging from *arithmetic and algebraic manipulation* to *geometry, function understanding* and *data interpretation*. Items are a mix of *multiple-choice* and *open-ended* questions with careful answer normalization for numeric/string responses [32]. The consolidation draws from chart- and figure-centric corpora (e.g., ChartQA, PlotQA, FigureQA) to ensure breadth in visual formats [122], [123], [124].

#### 3) WHAT IT MEASURES (BEYOND ACCURACY)

By requiring models to jointly parse *visual structures* (axes, legends, spatial relations) and perform *symbolic/numeric* operations, MathVista probes: (i) chart/diagram *comprehension*, (ii) *multi-step quantitative reasoning* grounded in images, and (iii) *robust answer extraction/normalization* for open-ended math responses [32]. These properties complement the generalist image–text benchmarks that underweight numeric competence.

#### 4) REPRESENTATIVE FINDINGS

Frontier MLLMs substantially underperform humans. While the original paper reported early multimodal models in the mid-30% range, recent leaders like GPT-4V have reached approximately *50%* accuracy, yet still trail human performance (around *60%*+), with persistent gaps in *function understanding* and *diagram/figure parsing* [32]. The results align with other visual-math probes (e.g., *PolyMATH*) that similarly expose brittle quantitative reasoning in current systems [125].

#### 5) WHAT IT LACKS (AS AN EVALUATION)

- *Limited multi-metric/process reporting.* MathVista primarily reports task accuracy; it does not natively cover calibration, robustness-to-perturbations, or efficiency (in contrast with HELM-style multi-metric reporting) [1].
- *Prompt/format sensitivity for open-ended scoring.* Numeric/string extraction in open-ended grading can be brittle to formatting and prompt templates; calibration and prompt-robust protocols are advisable [126], [127].
- *Static, non-adversarial refresh.* As a consolidated but largely *static* suite, it may saturate or accrue a train-test overlap; living/adversarial collection pipelines help sustain headrooms [13], [53], [116].
- *Contamination/memorization risk.* Merged content from widely circulated sources increases the overlap risk with pretraining corpora; audits and hidden splits are important for construct validity [37].
- *Narrow modality band.* The focus is on image + text; audio/video quantitative reasoning (e.g., reading values from narrated plots in videos) is beyond the scope of this study and can be complemented by other suites.

Key properties of the MathVista benchmark, designed to assess mathematical reasoning and visual problem-solving in multimodal systems, are detailed in Table 16.

### D. MM-VET: INTEGRATED-CAPABILITY, OPEN-ENDED EVALUATION

Many benchmarks test skills in isolation (e.g., only OCR or only object detection). *MM-Vet* shifts the focus to *integrated capabilities*, diagnosing whether models can chain these skills together to solve open-ended, composite problems.

#### 1) WHAT IT IS

*MM-Vet* is a diagnostic benchmark that emphasizes *integrated* vision–language capabilities with *open-ended* answers. Instead of isolating single skills, each item is constructed to require combinations of six core capabilities—*recognition, OCR, knowledge, language generation, spatial awareness* and *math*—and is graded with a large language model (LLM)-as-judge for unified scoring across answer styles [33]. The initial release comprised *200* images and *218* QA samples spanning *16* task types, with code and an online evaluator provided [33], [128], [129].

#### 2) COVERAGE AND DESIGN

Items are curated so that solving them requires *capability integration* (e.g., recognizing objects, reading scene text, localizing the referred region, and computing a result). Prompts elicit free-form responses rather than constrained choices in order to better reflect realistic usage. An LLM-based evaluator (default: *GPT-4*) assigns a normalized score (0–1) based on soft-matching criteria, enabling a single metric across heterogeneous questions/answers. Sanity checks (e.g., keyword matching) and ablations on the evaluator reliability have been explicitly reported [33], [130].

#### 3) WHAT IT MEASURES (BEYOND ACCURACY)

MM-Vet explicitly breaks out *core* vs. *integrated* skills and reports per-capability and per-integration slices, highlighting where models fail to compose abilities (e.g., OCR + Math, Recognition + Spatial). It also analyzes *system paradigms* (end-to-end LVLMs vs. tool-augmented agents) and *evaluator reliability* (LLM-as-judge vs. humans), offering insights that single-score, single-skill benchmarks miss [33].

#### 4) REPRESENTATIVE FINDINGS

(1) Frontier models continue to push the ceiling: while early leaders such as *GPT-4V* achieved *67.7%*, recent evaluations show *Claude 3.5 Sonnet* reaching *71.8%* and *GPT-4o* at *71.0%* [33], [114]. Despite this, complex integrated tasks (e.g., Spatial+Math) remain a bottleneck. (2) Tool-using agents (e.g., MM-ReAct [131]) can outperform end-to-end models on math- and knowledge-heavy integrations, suggesting benefits from explicit tool use [33]. (3) Upgrading the underlying LLM generally lifts integrated-capability scores, confirming that reasoning bottlenecks are often linguistic rather than purely visual [33].

**TABLE 17.** MM-Vet at a glance.

| Property | Details |
| --- | --- |
| Modalities | Image + Text (open-ended responses) |
| Scale | 200 images; 218 QA items; 16 task types |
| Core abilities | Recognition, OCR, Knowledge, Language Generation, Spatial Awareness, Math |
| Task design | Integrated capability compositions (pairwise and higher-order) |
| Metric | LLM-as-Judge Unified Score (0–1; GPT-4 evaluator) |
| Diagnostic focus | Capability *integration* (e.g., OCR+Math, Rec+Spatial) rather than single skills |
| Resources | Code/data and online evaluator available |

#### 5) WHAT IT LACKS (AS AN EVALUATION)

- *Small scale/statistical power.* The core set (200 images/218 items) limits per-slice confidence intervals and saturation analysis; larger, regularly refreshed sets would strengthen conclusions [33].
- *Judge dependence/sensitivity.* Reliance on a specific LLM-as-judge (GPT-4) raises concerns about prompt/rubric sensitivity, verbosity bias, and cost; standardized rubrics and human-agreement reporting are essential for validity [130], [132].
- *Modality scope.* Image–text only; no explicit coverage of temporal (video) or audio grounding, which are key failure modes surfaced by video-centric suites [35].
- *Process transparency.* No gold *rationales*/step annotations (contrast with VCR-Bench)—therefore it diagnoses *what* fails (capability integration) more than *where* in the reasoning chain it fails.
- *Multi-metric breadth.* The unified open-ended score does not natively cover calibration, robustness to prompt format, fairness/toxicity, or efficiency (cf. HELM-style multi-metric reporting) [1].
- *Contamination auditing.* As with most public benchmarks, explicit pretraining-overlap audits or hidden refresh splits are needed to preserve construct validity over time [133].

The design and evaluation metrics of the MM-Vet benchmark, which focuses on evaluating general multimodal reasoning and robustness, are summarized in Table 17.

### E. HALLUSIONBENCH: SYSTEMATIC BENCHMARKING OF MULTIMODAL HALLUCINATION

High accuracy on standard VQA often masks a model's tendency to fabricate information. *HallusionBench* serves as a critical negative control, explicitly testing for *hallucination* by presenting plausible but false claims that models must reject.

#### 1) WHAT IT IS

*HallusionBench* is a diagnostic benchmark that targets *hallucination* in large vision–language models (LVLMs). It constructs minimal, controlled image–question pairs—including edited images—to elicit and measure visual or knowledge-driven *illusions* in a yes/no QA setting [34]. The release reports scale (∼1129 questions over 346 images) and a taxonomy spanning *vision-dependent* vs. *vision-supplement*

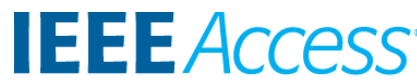

(knowledge) hallucinations, with subject categories (e.g., architecture, people, animals) and a small subset of consecutive-frame items (video-like) [34].

#### 2) COVERAGE AND DESIGN

Each item pairs a visual scene (real or lightly edited) with a claim that is *semantically precise* and verifiable from pixels and/or common knowledge. Questions are grouped by: (i) *illusion source* (vision-dependent vs. vision-supplement), (ii) *subject* (architecture, person, animal, object, etc.), and (iii) *modality format* (single image with some image cases) [34]. The yes/no format reduces parsing noise and allows the benchmark to use *contradictory* pairs (true and falsified variants) to test for consistency.

#### 3) WHAT IT MEASURES

Primary reports include *question-pair accuracy* (joint correctness across paired claims) and per-category breakdowns; consistency across contradictory pairs surfaces whether models both affirm truths and *reject* plausible-sounding false statements, a hallmark of hallucination robustness [34]. Because the items are tightly controlled, errors can be easily attributed to perception vs. knowledge gaps.

#### 4) REPRESENTATIVE FINDINGS

Despite strong headline performance elsewhere, frontier LVLMs perform poorly in HallusionBench. The authors report an overall *question-pair accuracy* of approximately *32%* (Vision-Dependent subset: ∼28%) for early GPT-4V versions, indicating frequent failures to reject subtly false claims even when the corresponding true variant is answered correctly [34]. Error analyses showed both vision-dependent slips (mislocalized or fabricated objects) and vision-supplement mistakes (commonsense/knowledge intrusions), underscoring that hallucinations are *not* just OCR or captioning artifacts.

#### 5) WHAT IT LACKS (AS AN EVALUATION)

- *Binary answers only.* Yes/no simplifies scoring but does not capture *rationale quality*, calibration, or partial credit; open-ended or CoT-graded variants would add process visibility (cf. VCR-Bench for rationale scoring) [113].
- *Limited process metrics.* No native measures for calibration, efficiency, or bias/toxicity; multi-metric reporting (e.g., HELM-style) is outside the current scope [1].
- *Narrow temporal coverage.* Only a small discriminative image subset, broader video reasoning stress (actions, long-range context), is better covered by video suites (e.g., Video-MME, SEED-Bench) [35], [134].
- *Static exposure risk.* Public test items can be memorized; contamination audits or hidden test servers mitigate leakage but are not intrinsic to the dataset [37], [133].
- *Judging and preference gaps.* Hallucination here is scored against ground truth, not human preferences; complementary human-arena setups (LVLM-eHub) can reveal verbosity and style biases that static metrics miss [115], [135].

**TABLE 18.** HallusionBench at a glance.

| Property | Detail |
|---|---|
| Modalities | Image (with some consecutive-image cases) + Text |
| Scale | ∼1129 Qs, 346 images (incl. edited images) |
| Task | Yes/No verification with paired (true vs. false) claims |
| Metrics | Question-Pair Accuracy (Consistency); Yes/No Exact Match |
| Diagnostic focus | Vision-dependent vs. vision-supplement hallucination; consistency |

The HallusionBench benchmark, which evaluates language hallucination and visual illusion phenomena in large vision-language models, is summarized in Table 18.

### F. MMMU: MASSIVE MULTI-DISCIPLINE MULTIMODAL UNDERSTANDING

Moving beyond daily commonsense, *MMMU* pushes the difficulty ceiling to the expert level. It functions as a domain-specific "qualifying exam," requiring college-level knowledge in Science, Technology, Engineering, and Mathematics (STEM) and professional fields to diagnose specialized reasoning gaps.

#### 1) WHAT IT IS

*MMMU* is a college/expert-level exam for MLLMs with *11,500* questions across diverse disciplines (e.g., Art & Design, Health & Medicine, Tech & Engineering), using heterogeneous visuals such as diagrams, charts, maps, and formulas [21].

#### 2) COVERAGE AND DESIGN

MMMU aggregates real-world, figure-centric problems from exams, textbooks, and professional materials across the STEM and non-STEM fields. Items interleave *image* and *text* and mix *multiple-choices* with *open-ended* formats. Visuals cover plots/tables, geometric diagrams, scientific schematics, and scanned pages; prompts are authored/selected to require *grounding in the visual artifact* rather than text-only cues. The evaluation toolkit provides an *answer normalization* for numeric/string responses and standardized templates thus results are comparable across disciplines. Reports typically include *per-discipline* breakdowns and optional toggles (e.g., with/without auxiliary tools) [2], [21].

#### 3) WHAT IT MEASURES

MMMU targets *expert domain knowledge* and *diagram/figure comprehension* under interleaved image–text inputs, with both MCQ and open-ended formats. It stresses cross-modal grounding on scientific figures and real exam materials, and reports per-discipline breakdowns to profile strengths/weaknesses [2], [21].

#### 4) REPRESENTATIVE FINDINGS

While the original paper reported *55.7%* accuracy for early GPT-4V versions, the frontier has advanced rapidly. Recent evaluations show *GPT-4o* reaching *69.1%* and *Gemini 1.5 Pro* at *65.8%*, approaching but not yet matching human expert performance (∼90%) [21], [136].

**TABLE 19.** **MMMU at a glance.**

| Property | Detail |
|---|---|
| Scale / Domains | 11500 Qs across multiple college/expert disciplines |
| Modalities | Image + Text (heterogeneous: charts, maps, formulas) |
| Tasks | Multiple-choice and open-ended QA |
| Primary metrics | Micro-averaged Accuracy (0-shot); LLM-extracted |
| Diagnostic focus | Expert knowledge, diagram parsing, deliberate reasoning |

#### 5) WHAT IT LACKS

- *Process supervision.* MMMU scores outcomes, not intermediate reasoning, and lacks step-tagged rationales (e.g., perception vs. reasoning steps) that enable process diagnostics [113], [137].
- *Open-ended scoring reliability.* For free-form answers, exact-match or LLM-as-judge protocols can misalign with human judgments and be sensitive to rubric/prompting, motivating stricter judge designs [130], [132], [138], [139].
- *Robustness checks.* MMMU does not natively include choice-order robustness such as *CircularEval* from MMBench, which reduces the position bias in MCQ [17].
- *Modal scope.* It is primarily image–text; long-horizon video/audio temporal reasoning is beyond the scope of video-centric suites [35], [134], [140].
- *Living/adversarial refreshment.* MMMU is a static set; it lacks human-in-the-loop adversarial rounds or recency tracks that sustain difficulty and reduce overfitting [13], [39].
- *Contamination risk.* As with public benchmarks that reuse exam-like content, training–test memorization can inflate scores without auditing [37], [133].
- *Specialized STEM depth.* Some competencies (e.g., multi-step mathematical reasoning over formal notation) are better stress-tested by targeted suites such as PolyMATH [125].

Key properties and evaluation domains of the MMMU benchmark, which targets multi-modal multi-task understanding, are outlined in Table 19.

### *G. MMBENCH: FINE-GRAINED ABILITY PROFILING FOR LVLMs*

A major threat to validity in multiple-choice exams is position bias (e.g., models preferring option A). *MMBench* addresses this *reliability* axis by introducing CircularEval, forcing models to demonstrate consistency across permuted options.

#### 1) WHAT IT IS

*MMBench* is a broad image–text benchmark that organizes evaluation into a hierarchical taxonomy of *20* ability dimensions under *Perception* and *Reasoning*, producing an ability *profile* rather than a single headline score [17], [114]. A key reliability feature is *CircularEval*, which permutes multiple-choice options to suppress the position bias and fragile pattern-matching inherent in standard VQA.

#### 2) COVERAGE AND DESIGN

Items are curated as multiple-choice questions that probe fine-grained skills across *20* dimensions (e.g., OCR/text reading, localization, attribute recognition, relation and logic reasoning). The evaluation adopts *choice-permutation* (*CircularEval*) to test invariance to answer ordering and a standardized *answer-extraction* toolkit (regex + LLM-assisted parsing) to harmonize scoring across models and prompt templates [114]. Public releases (via OpenCompass) provide configs, leaderboards, and analysis tools for reproducible runs and per-dimension reporting [17], [114].

#### 3) WHAT IT MEASURES

MMBench probes OCR/text reading, object/attribute recognition, localization, part–whole and relational reasoning, and basic logical inference through carefully curated MCQs. Evaluation reports (i) *CircularEval* accuracy, (ii) per-dimension breakdowns (ability radar), and (iii) robust choice extraction (regex + LLM-assisted parsing) for fair cross-model comparison [17], [114]. Because abilities are disaggregated, MMBench can surface capability trade-offs that single-score suites often hide, complementing multi-metric philosophies such as the Holistic Evaluation of Language Models (HELM) [1].

#### 4) REPRESENTATIVE FINDINGS

Current frontier models such as *GPT-4o* and *Gemini 1.5 Pro* achieve ∼ *85%* accuracy on the leaderboard (*84.8%* and *84.4%* respectively), demonstrating strong generalist capabilities [10], [114]. However, across model families, *CircularEval* typically lowers headline scores by *10–20%*, revealing sensitivity to superficial cues (e.g., choice order) and improving reliability [17]. Fine-grained perception tasks—OCR on small text, counting in clutter, and precise spatial relationships—remain challenging for many open models, consistent with targeted diagnostics (e.g., CountQA's numerosity stress [111]).

#### 5) WHAT IT LACKS

- *Process supervision.* MMBench scores *outcomes* (answers), not intermediate *reasoning steps*; it lacks step-tagged rationales (e.g., perception vs. reasoning tags) that enable process diagnostics such as VCR-Bench (video CoT) and GeoChain (stepwise geo-reasoning) [113], [137].
- *Judge the stability of open-ended extraction.* When LLMs are used to extract/select choices, results can be sensitive to judge prompts/rubrics; best practice is to control LLM-as-judge bias and report agreement per [130], [132], [139].
- *Beyond choice-order robustness.* CircularEval addresses position bias, but other robustness axes, prompt/format

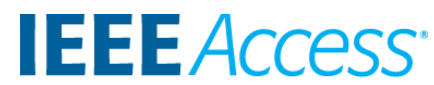

**TABLE 20.** **MMBench at a glance.**

| Property | Detail |
|---|---|
| Modalities | Image + Text |
| Ability coverage | 20 dimensions (Perception & Reasoning) |
| Tasks | Multiple-choice QA with choice permutation (*CircularEval*) |
| Answer extraction | Regex + LLM-assisted parsers (standardized) |
| Metrics | CircularEval Accuracy (Robustness-weighted via choice permutation) |
| Diagnostic focus | OCR, fine-grained perception, attributes, relations, logic |
| Known gaps (external) | Process rationales, negation stress, counting depth, video/audio, adversarial refresh |

sensitivity and decoding settings-still affect scores, considering contextual calibration and prompt controls [126], [127].

- *Negation and linguistic logic.* MMBench includes logic items, but explicit *negation* stress tests (NegBench/NegVQA) often reveal larger gaps than general MCQs [105], [106].
- *Counting and compositional binding depth.* Dedicated stressors (CountQA, COLA, Winoground) provide harder and more controlled probes of numerosity and attribute–relation binding than general-purpose MCQs [84], [111], [141].
- *Modal scope.* MMBench is *image–text*; it does not cover long-horizon video/audio temporal reasoning, which is better assessed by SEED-Bench (image+video) and Video-MME (video+subtitles+audio) [35], [134], [140].
- *Living/adversarial refreshment.* It is a static suite; human-in-the-loop adversarial collection and recency refresh (e.g., Dynabench and, RealTimeQA) can sustain difficulty and reduce overfitting [13], [39].
- *Contamination auditing.* Public MCQs risk leakage/memorization in web-scale pretraining; audits and hidden/rotating test shards mitigate this [37], [133].
- *Hallucination analysis.* Object hallucination is not a first-class axis here; complementary arenas that analyze hallucination and human preferences (e.g., LVLM-eHub) help align scores with user judgments [115], [142].

MMBench's category coverage and scoring protocol are summarized in Table 20.

### *H. SEED-BENCH: UNIFIED SPATIAL (IMAGE) AND TEMPORAL (VIDEO) REASONING*

To unify the evaluation of spatial and temporal understanding, *SEED-Bench* provides a standardized protocol across both image and video modalities, allowing for a direct comparison of a model's static perception versus its dynamic reasoning.

#### 1) WHAT IT IS

*SEED-Bench* is a large-scale suite for *both* image and video understanding that provides *19k+* multiple-choice questions across *12* evaluation dimensions spanning spatial (e.g., localization, relations, counting) and temporal (e.g., action recognition/prediction, procedure understanding) competencies [134], [143], [144]. Items are templated MCQs with carefully designed distractors to limit shallow priors and emphasize cross-frame reasoning in the video track [134].

#### 2) WHAT IT MEASURES

SEED-Bench stresses (i) *fine-grained spatial understanding* (attributes, part–whole, precise relations), (ii) *temporal reasoning* (recognizing and *predicting* actions, ordering procedures), and (iii) *basic numerosity* by counting in cluttered scenes under a uniform MCQ protocol enabling *per-dimension* accuracy breakdowns and *per-modality* (image vs. video) slices [134]. Because the suite mixes the spatial and temporal axes, it helps separate perception-only gains from genuine time-aware reasoning, complementing broader surveys of multimodal evaluation [2].

#### 3) REPRESENTATIVE FINDINGS

Across open and closed models, accuracy on temporal dimensions typically trails over spatial ones; models degrade on action prediction and multi-step procedure understanding relative to single-frame perception tasks [134]. Counting and tight spatial relations also remain weak spots for many LVLMs, consistent with focused diagnostics such as CountQA (numerosity in the wild) [111]. More generally, the literature shows that video understanding becomes markedly harder as temporal horizons grow and low-level cues (motion/ASR) are required [140]; long-horizon benchmarks with audio/subtitles (e.g., Video-MME) likewise report steep drops as clip length increases [35].

#### 4) WHAT IT LACKS

- *Long-horizon audio-aware evaluations.* SEED-Bench videos are short clips and vision-only; it does not probe speech/*subtitles* or longer (minutes–hour) contexts where audio–text fusion and memory matter [35], [140].
- *Process-level scoring.* The suite evaluates outcomes (MCQ answers) but not the *reasoning process*. Step-tagged rationales and chain-of-thought comparisons—for example, VCR-Bench's perception vs. reasoning tags or GeoChain's stepwise geo-reasoning— enable finer diagnosis of where models fail [113], [137].
- *Targeted logic stressors.* While SEED-Bench covers general logic, it lacks specialized probes for *negation* and compositional binding that expose affirmation bias and attribute–relation swaps (e.g., NegBench/NegVQA, COLA, Winoground) [84], [105], [106], [141], [145].
- *Living/adversarial refreshment.* It is a static suite; human-in-the-loop adversarial rounds and periodic refresh (Dynabench, RealTimeQA) help sustain the difficulty and mitigate overfitting to public test sets [13], [39].
- *Contamination auditing.* Public MCQs risk pretraining leakage, and contamination audits and hidden/rotating shards are recommended to preserve construct validity [37], [133].

**TABLE 21. SEED-bench at a glance.**

| Property | Detail |
|---|---|
| Modalities | Image + Video (text prompts; vision-only clips) |
| Scale | 19k+ MCQs; 12 evaluation dimensions |
| Metrics | Accuracy (Multiple Choice; 4-option) |
| Diagnostic focus | Spatial relations, counting, action recognition/prediction, procedure understanding |
| Known gaps (external) | Long-horizon audio/subtitle fusion; process-level scoring; negation/compositional stress; living/adversarial refresh; contamination audits; human preference/hallucination axes |

- *Human preference & hallucination analysis.* SEED-Bench does not include human *arena* comparisons or object-hallucination axes; complementary hubs (LVLM-eHub) and multi-metric frameworks (HELM) cover these aspects [1], [115], [142].

The SEED-Bench task taxonomy and scoring are listed in Table 21.

### *I. GeoChain: CHAIN-OF-THOUGHT FOR GEOGRAPHIC REASONING*

Geographic localization requires fusing subtle visual cues with world knowledge. *GeoChain* evaluates this via *Chain-of-Thought* (CoT), diagnosing the specific step—perception or reasoning—where the localization process fails.

#### 1) WHAT IT IS

*GeoChain* is a Chain-of-Thought (CoT)–style benchmark for *stepwise* geo-localization and geographic reasoning over street-level imagery. Each item guides models from coarse regional cues (continent/country) toward fine-grained localization (city/landmark) with interleaved visual, spatial, and cultural signals [137].

#### 2) COVERAGE AND DESIGN

Items are constructed as structured *21-step* chains that progress from *coarse* (continent/hemisphere/region) to *fine* (country → city/landmark) hypotheses [137]. Street-level scenes are sampled to span diverse geographies and conditions (urban/rural; varied climate zones; day/night; different weather) and to expose informative cues such as scripts and languages on signage, road markings and driving side, vegetation/biome, architectural styles, and cultural artifacts [137]. At each hop, the model must (i) extract salient *perceptual* evidence, (ii) commit to a location hypothesis, and (iii) optionally justify that hypothesis using a brief CoT statement [137]. Distractors are designed as *plausible confounders* (e.g., Spanish vs. Portuguese signage; tropical vs. Mediterranean flora) so that success requires fusing multiple weak cues rather than exploiting a single shortcut, which mirrors long-standing practices in photo geolocalization [146], [147].

#### 3) WHAT IT MEASURES

GeoChain explicitly probes (i) *hierarchical spatial reasoning*—from broad region recognition down to local place inference, (ii) *evidence decomposition* via multi-step CoT, and (iii) *the use of visual/cultural priors* (e.g., signage styles, vegetation, driving side) that have long been central to image geolocalization [146], [147]. By requiring intermediate justifications, it helps separate *perception* (extracting visual cues) from *reasoning* (mapping cues to geographic hypotheses), akin to the recent rationale-tagged evaluations in video QA [113].

#### 4) REPRESENTATIVE FINDINGS

The reported results show that many MLLMs stumble on *intermediate* CoT hops—for example, correctly reading a language/script yet failing to integrate it with vegetation/climate cues leading to near-miss finals (right region, wrong country) [137]. This mirrors the classic geolocalization challenges where coarse cues are easy, but fine localization requires fusing *multiple weak signals* [146], [147].

#### 5) WHAT IT LACKS

- *Distance-based scoring.* GeoChain emphasizes stepwise accuracies; adding continuous geodesic metrics (median/mean great-circle error) would align with standard photo geolocalization practices (e.g., IM2GPS, PlaNet) and reveal near-miss progress [146], [147].
- *Cross-view/map grounding.* Many real systems reason with aerial/maps vs. ground-view matching, and incorporating cross-view settings would stress a distinct axis of spatial reasoning not covered by street-level only [146].
- *OCR/multilingual signage stress.* Fine localization often hinges on reading storefronts/road signs; explicit multilingual OCR drills (as profiled in OCR-centric ability axes such as MMBench) could sharpen the diagnosis [17].
- *Temporal/embodied context.* Single-image inference omits motion cues and interactions. Longer-horizon video (e.g., Video-MME) and embodied navigation tie-ins (ALFRED/Habitat) would test whether models can plan and refine hypotheses over time and action [35], [42], [117], [118], [119], [120].
- *Process scoring beyond accuracy.* Beyond per-step correctness, CoT quality metrics (precision/recall over rationale steps as in VCR-Bench) and HELM-style multi-metric reporting (calibration, robustness, bias) would strengthen construct validity [1], [113].
- *Living/adversarial refresh and contamination audits.* Public, niche imagery can be leaked into pretraining. Human-in-the-loop adversarial rounds (Dynabench) and contamination auditing are recommended to preserve difficulty and measurement integrity [13], [37], [133].

The design and evaluation protocol of the GeoChain benchmark, which focuses on geographic and spatial reasoning, are summarized in Table 22.

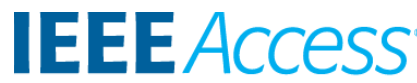

**TABLE 22.** GeoChain at a glance.

| Property | Detail |
|---|---|
| Modalities | Image + Text (stepwise prompts with CoT) |
| Task | Hierarchical geo-localization from region → city/landmark |
| Metrics | Per-step Accuracy; Overall Pass Score [137] |
| Suggested metrics | Geodesic error (km); CoT precision/recall; HELM-style multi-metric reporting |
| Diagnostic focus | Visual grounding vs. spatial/cultural reasoning; fusion of weak cues |
| Known gaps (external) | Cross-view/map grounding; OCR multilingual stress; temporal/embodied context; living/adversarial refresh; contamination audits |

**TABLE 23.** VCR-Bench at a glance.

| Property | Detail |
|---|---|
| Modalities | Video + Text (open-ended QA + CoT rationale) |
| Scale | ∼1034 QA items with human CoT; step tags: perception vs. reasoning |
| Tasks | Open-ended QA; rationale generation and stepwise alignment to gold CoT |
| Metrics | CoT Score (Stepwise Precision/Recall); Answer Accuracy |
| Diagnostic focus | Temporal–spatial perception vs. logical inference; process fidelity |
| Known gaps (external) | Audio/subtitles, scale/breadth, judge/prompt sensitivity, prompt/decoding variance, contamination, multi-metric breadth |

### J. VCR-BENCH: VIDEO CHAIN-OF-THOUGHT WITH PERCEPTION VS. REASONING TAGS

Final answers can be right for the wrong reasons. *VCR-Bench* targets the *process fidelity* of video understanding, requiring models to generate rationale tags that distinguish between what they see (perception) and what they infer (reasoning).

#### 1) WHAT IT IS

*VCR-Bench* is a video-centric benchmark that evaluates the reasoning *process* and, not just the final answers. Each item pairs a video QA with a human-authored stepwise Chain-of-Thought (CoT) whose steps are tagged as *perception* vs. *reasoning*, extending the spirit of image-based VCR to multi-frame settings [113], [148].

#### 2) COVERAGE AND DESIGN

Short multi-frame clips with open-ended questions: Each QA has a minimal gold rationale decomposed into ordered steps with tags (perception vs. reasoning). Items are chosen to discourage single-frame shortcuts and require temporal–spatial integration [113].

#### 3) WHAT IT MEASURES

Answer *accuracy* and a *CoT Score* that aligns model rationales to gold and computes stepwise precision/recall; per-tag breakdowns localize failures to perception versus reasoning [113].

#### 4) REPRESENTATIVE FINDINGS

Accuracy often masks process deficits: strong MLLMs (such as GPT-4o and Gemini 1.5 Pro) consistently score lower on *perception* steps than on *reasoning* steps, revealing that the primary bottleneck is often extracting correct visual evidence rather than the subsequent logic [113].

#### 5) WHAT IT LACKS

- *Modality coverage.* Limited treatment of subtitles/audio vs. video suites, such as Video-MME [35].
- *Scale and breadth.* ∼1034 QAs, narrower coverage than general video/image–video suites (e.g., SEED-Bench) [134].
- *Judge/prompt sensitivity.* CoT alignment and any LLM-as-judge use can be prompt/rubric sensitive, reporting agreement, and control biases [130], [132], [139].
- *Prompt/decoding variance.* The CoT outcomes swing with template/decoding; calibrated prompting and multi-template aggregation improve stability [126], [127], [149].
- *Contamination risk.* Public items risk leakage: Audits/hidden splits mitigate memorization [37], [133].
- *Multi-metric breadth.* Missing calibration/robustness/efficiency reporting (cf. HELM) [1].

The core components and evaluation setup of the VCR-Bench benchmark are summarized in Table 23, highlighting its tasks, rationale structure, and metrics for commonsense reasoning.

### K. COMPARATIVE LANDSCAPE

Table 24 consolidates the *broad-spectrum exams* (e.g., MMMU, MMBench, SEED-Bench) with *deep-dive diagnostics* (e.g., MathVista, HallusionBench, MM-Vet, GeoChain, VCR-Bench) and adds a long-horizon, audio-aware video suite (Video-MME). We avoid repeating per-benchmark details already covered in Section V-B–Section V-J; instead, we emphasize contrasts in *modalities, task formats, scoring methods* and *unique diagnostic focus*.

Use *MMMU* for breadth at the expert level; *MMBench* for fine-grained *ability profiles* with robustness to choice order. *SEED-Bench* for unified spatial vs. temporal slices; *Video-MME* when long-horizon memory and audio/subtitle fusion matter; *MathVista* to stress visual *quantitative* reasoning; *MM-Vet* to assess *integrated* capability compositions under open-ended grading; *HallusionBench* to isolate hallucinations (rejecting false claims), *GeoChain* to examine stepwise CoT for geo-localization, *VCR-Bench* to evaluate *process fidelity* in video, and legacy *GQA/VCR* for historically important constructs (compositionality, answer+justification).

#### 1) PROCESS-FIRST EVALUATION

Echoing Section V-J and Section V-I, a notable trend is scoring the *reasoning process* (not just the outcomes). *VCR-Bench* adds step-tagged CoT with precision/recall over the rationale steps, whereas *GeoChain* scores intermediate hops in hierarchical localization. These complement outcome-centric suites and help diagnose whether models are "right for the right reasons." Open-ended tasks that still rely on LLM-as-judge (*MM-Vet*, parts of MMMU), report

**TABLE 24.** A comparative analysis of modern multimodal reasoning benchmarks.

| Benchmark (Year) | Primary Modalities | Task Format | Evaluation Method | Key Reasoning Skills Tested | Unique Contribution / Diagnostic Focus |
|---|---|---|---|---|---|
| *MMMU* (2023) [21] | Image + Text (heterogeneous) | MCQ, Open-Ended QA | Micro-averaged Accuracy (0-shot); LLM-extracted | Expert domain knowledge; deliberate reasoning; figure/diagram parsing | College-/expert-level multi-discipline exam using real exam materials. |
| *MMBench* (2023) [17] | Image + Text | MCQ with choice permutation (*CircularEval*) | CircularEval Accuracy (Robustness-weighted) | Fine-grained perception (OCR, localization), attributes/relations, logic | Ability *profile* over 20 dimensions; robust to position-bias. |
| *SEED-Bench* (2023) [134], [143] | Image + Text; Video + Text | MCQ (templated, curated distractors) | Accuracy (Multiple Choice; 4-option) | Spatial relations, counting; temporal actions/prediction; procedures | Unified spatial (image) + temporal (video) reasoning under a single protocol. |
| *Video-MME* (2024/2025) [35], [112] | Video + Subtitles + Audio + Text | MCQ (3 per video) with modality toggles | Accuracy (MCQ; 4-option; 0-shot) | Long-horizon temporal memory; audio-/subtitle-grounded reasoning | Full-spectrum video suite (900 videos, 256h) probing fusion beyond frames. |
| *MathVista* (2023) [32] | Image + Text (charts, diagrams, tables, scanned pages) | MCQ and Open-Ended QA (normalized) | Accuracy (Normalized Numeric/String Exact Match) | Visual parsing + quantitative reasoning (functions, geometry, data interpretation) | Consolidates 28 visual-math sets; adds IQTest/FunctionQA/PaperQA for breadth. |
| *MM-Vet* (2023) [33], [128], [129] | Image + Text | Open-Ended QA (integrated capabilities) | LLM-as-Judge Unified Score (0–1) | *Integration* of recognition, OCR, knowledge, language, spatial, math | Scores capability *compositions* (e.g., OCR+Math), not isolated skills. |
| *HallusionBench* (2023) [34] | Image + Text (some consecutive images) | Yes/No verification (paired true/false claims) | Question-Pair Accuracy (Consistency) | Hallucination robustness; rejecting plausible false claims | Controlled illusions (edited/precise claims) to isolate hallucination failure modes. |
| *GeoChain* (2025) [137] | Image + Text (stepwise prompts) | CoT QA sequence (hierarchical localization) | Per-step Accuracy; Overall Pass Score | Visual grounding vs. spatial/cultural reasoning; coarse→fine localization | Benchmarks the *process* of geo-reasoning via multi-step CoT. |
| *VCR-Bench* (2025) [113] | Video + Text | Open-Ended QA + CoT rationale (tagged steps) | CoT Score (Stepwise Precision/Recall) | Temporal–spatial perception vs. logical inference; causal understanding | Explicit *process* scoring with perception vs. reasoning tags. |
| *GQA* (2019) [26] | Image + Text | Open-Ended QA | Accuracy; consistency/validity/plausibility; grounding | Compositional reasoning; multi-step inference; spatial logic | Early large-scale *reasoning* suite with diagnostic metrics beyond accuracy. |
| *VCR* (2019) [28] | Image + Text | MCQ + Rationale selection (Q→AR) | Joint Answer+Rationale accuracy | Visual commonsense; intents/goals; causal/explanatory reasoning | First to *require justification*, pushing beyond answer-only evaluation. |

prompt/rubric controls and human-agreement to mitigate judge sensitivity [130], [132], [139].

### 2) GAPS AND COMPLEMENTARITY (BRIEF)

No single suite covered everything. A long-horizon audio-aware video (*Video-MME*) complements short-clip vision-only video in *SEED-Bench*, *MathVista* fills numeric reasoning gaps underweighted elsewhere, *HallusionBench* targets hallucination explicitly, and *MMBench* offers robustness-minded MCQ scoring (*CircularEval*) that many sets lack. Multi-metric transparency (HELM-style) and contamination audits remain broadly useful overlays across all suites [1], [37], [133].

*MMMU (2023).* A broad "expert exam" spanning 11500 items across heterogeneous visuals and disciplines; stresses figure/diagram comprehension and domain knowledge. Useful when you need a cross-disciplinary bar beyond commonsense; see details in Section V-F [21].

*MMBench (2023).* Comprehensive "cognitive skills assessment." Instead of a single score, it returned an ability profile over 20 perception/reasoning dimensions. Its *CircularEval* permutes the choice order to suppress position bias, typically lowering headline accuracy by 10–20% vs. naive MCQ evaluation; see Section V-G [17], [114].

*SEED-Bench (2023).* A large "aptitude test" for both spatial (image) and temporal (video) competencies, with 19k+ MCQs over 12 dimensions. Balanced design makes it easy to compare perception versus time-aware reasoning (see Section V-H [134], [143]).

*Video-MME (2024/2025).* Long-horizon, audio/subtitle-aware video QA (900 videos; 256h; 3 MCQs per video) with S/M/L duration splits and modality toggles. Ideal when you need to probe memory over minutes and fusion beyond frames; see Section V-B [35], [112].

*MathVista (2023).* Unified visual mathematical reasoning suite that consolidates 28 datasets and adds IQTest/FunctionQA/PaperQA. Targets chart/diagram parsing plus symbolic/numeric reasoning; see Section V-C [32].

*MM-Vet (2023).* Open-ended *integrated capability* evaluation (recognition, OCR, knowledge, language generation, spatial, math) with an LLM-as-judge for unified scoring. Answers whether models can *compose* skills (e.g., OCR + Math); see Section V-D [33], [128], [129].

*HallusionBench (2023).* Minimal, controlled image+claim pairs (including edited images) to elicit and measure hallucinations. Binary yes/no with paired true/false claims highlights the failure to *reject* plausible falsehoods ( Section V-E [34]).

*GeoChain (2025).* Chain-of-Thought (CoT) for stepwise geo-localization: coarse → fine spatial reasoning over street-level imagery with per-step scoring. Useful to diagnose where reasoning breaks (perception vs. spatial logic); see Section V-I [137].

*VCR-Bench (2025).* Video QA with human CoT rationales whose steps are tagged as *perception* versus *reasoning*; reports CoT precision/recall in addition to accuracy. Measures whether models are right for the *right reasons*; see Section V-J [113], [148].

*GQA (2019) and VCR (2019).* Legacy foundations: GQA introduced programmatic compositional QA with diagnostic metrics; VCR required answer *and* rationale selection, pushing beyond answer-only evaluations. We cite these for the historical context and design patterns [26], [28].

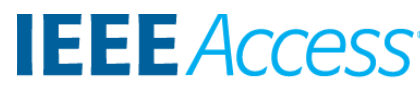

### L. SYNTHESIS OF LEVEL III: THE REASONING ERA

The Reasoning Era represents the current frontier, where MLLMs are evaluated on their ability to perform expert-level tasks across diverse modalities. Table 25 captures the fundamental paradox of this era: while *Skill* depth has reached college-level proficiency (MMMU [21]), it has come at the direct expense of *Hygiene* and *Reliability*. The ''Train-on-the-Web'' paradigm has created a Hygiene Crisis where guaranteeing clean test sets is nearly impossible [133], while the shift to stochastic, open-ended generation (evaluated by LLMs) has made scoring less reproducible than the deterministic metrics of the Perception Era. Furthermore, the *Cost* of evaluation has risen exponentially [150]. As we move to *Level IV*, the field attempts to resolve this hygiene crisis through procedural simulation and embodiment, albeit at even greater computational cost.

## VI. LEVEL IV: ABSTRACT AND CREATIVE INTELLIGENCE (THE UNCHARTED TERRITORIES)

As models begin to perform strongly on complex, knowledge-based reasoning tasks with verifiable ground-truth answers, the frontier of evaluation is pushing into the more ambiguous, subjective, and dynamic domains that constitute hallmarks of human intelligence [152]. The next level of the AI cognitive examination moves away from static question-answering and toward assessing capabilities in interactive environments, understanding nuanced social dynamics, and generating novel and valuable artifacts. In these uncharted territories, objective, right-or-wrong metrics become less relevant, giving way to evaluations that are process-oriented, behavioral, and aligned with human judgment.

### A. BEYOND STATIC Q&A: THE NEXT FRONTIER OF EVALUATION

The limitations of static benchmarks are becoming increasingly apparent. A model that can answer questions about the picture of a kitchen is not necessarily capable of acting within one. True intelligence is not passive, but active, as demonstrated through goal-directed behavior in a dynamic world. This has led to a growing interest in evaluating AI systems on their ability to plan, interact, and create—tasks where the ''correct answer'' is not a single entity to be retrieved, but a successful sequence of actions or a high-quality, contextually appropriate output output.

### B. VirtualHome

#### 1) WHAT IT IS

A simulator and knowledge base that encodes everyday household activities as *programs*—ordered sequences of atomic actions over objects—paired with natural-language descriptions and synthetic videos with dense ground truth (pose, segmentation, depth) [43]. The programs are executable in a Unity3D home simulator [43].

#### 2) COVERAGE AND DESIGN

Crowdsourced *2,821* ActivityPrograms spanning *75* atomic actions and *308* objects; a Scratch-style interface with *77* block types; Unity3D homes averaging ∼357 object instances per scene [43]. *Scoring:* Longest Common Subsequence (LCS) similarity (0–1) and Execution Success Rate (SR) in the simulator [43].

#### 3) WHAT IT MEASURES

- *Procedure induction*: mapping language or video to executable programs [43].
- *Commonsense completion*: inferring omitted but necessary steps [43].
- *Causal state tracking*: preconditions/effects across multi-step routines [43].

#### 4) WHAT IT LACKS

- *Egocentric perception:* primarily third-person renderings; limited egocentric viewpoints (contrast with ALFRED [117]).
- *Physical realism:* simplified physics/control compared to robotics simulators such as AI2-THOR or Habitat [120], [153].
- *Failure patterns in practice:* missing critical intermediate steps, symbol grounding to the wrong instance, overfitting to frequent program templates [43].

### C. ALFRED (ACTION LEARNING FROM REALISTIC ENVIRONMENTS AND DIRECTIVES)

#### 1) WHAT IT IS

A vision–language benchmark in AI2-THOR, in which an egocentric agent follows high-level goals and a step-by-step language to complete long-horizon household tasks. Dataset: *25,743* directives for *8,055* expert demonstrations (avg. ∼50 steps) across *120* scenes; interaction via object-centric masks [117], [153].

#### 2) COVERAGE AND DESIGN

Seven task families (Pick&Place, Stack&Place, Pick-Two&Place, Clean&Place, Heat&Place, Cool&Place, Examine-in-Light) with train/val/test splits and *seen* vs. *unseen* scenes to probe generalization and mixed navigation–manipulation actions (pickup/put, open/close, toggle, slice, heat/cool, clean) [117]. *Scoring:* Task Success Rate (SR), Goal-Condition Success (GCS), and Path-weighted Success (SPL) to penalize inefficiency [117], [154].

#### 3) WHAT IT MEASURES

- Long-horizon grounding of language into perception–navigation–manipulation sequences [117].
- Subgoal compositionality and progress tracking [117].
- Generalization from seen to unseen scenes/objects [117].

**TABLE 25.** Framework synthesis: Level III (Expert Reasoning).

| Dimension | Rating | Justification and Evidence |
|---|---|---|
| *Skill* | ★★★★★ | *Expert-Level.* Reached college-level proficiency in STEM and professional domains. MMMU [21] and MathVista [32] test complex diagram parsing and multi-step logic, far exceeding the capabilities of previous eras. |
| *Reliability* | ★★ | *Low (Stochasticity).* Major regression from Level I. Open-ended evaluation relies on LLM-as-a-Judge, which suffers from verbosity bias and prompt sensitivity [130]. Even MCQs require expensive "CircularEval" to correct for position bias [17]. |
| *Robustness* | ★★★ | *Moderate.* Models are robust to perception noise but prone to *Hallucination*. HallusionBench reveals that SOTA models frequently fail to reject plausible but false claims [34]. |
| *Hygiene* | ★ | *Critical Crisis.* The "Train-on-the-Web" paradigm makes decontamination nearly impossible. Audits frequently detect N-gram overlaps between test sets and training corpora [2], [133]. |
| *Cost* | ★ | *Very High.* Generative inference is computationally expensive. Evaluating a single sample often requires decoding hundreds of tokens or processing long-context video (e.g., Video-MME [35]), limiting scalability. |
| *Fairness* | ★★★★ | *High Priority.* Unlike Level I, modern benchmarks explicitly incorporate "Red Teaming" and safety evaluations to detect toxicity and stereotype bias [1], [151]. |

#### 4) WHAT IT LACKS

- *Interaction brittleness:* object-centric masks can be sensitive to visual distribution shifts [117].
- *Credit assignment:* partial observability and long horizons strain memory/planning [42], [117].
- *Failure patterns in practice include* state-tracking errors (e.g., repeated heat/clean loops), navigation drift in unseen scenes, mis-grounded referring expressions, and inefficient paths that depress efficiency-weighted scores [117].

### D. MUEP (MULTIMODAL BENCHMARK FOR EMBODIED PLANNING)

#### 1) WHAT IT IS

An International Joint Conference on Artificial Intelligence 2024 (IJCAI-24) benchmark emphasizing multi-turn planning with multimodal observations, built primarily on ALFWorld (aligned with ALFRED). It offers fine-grained diagnostics and a large bank of expert demonstration episodes [155].

#### 2) COVERAGE AND DESIGN

*14,927* expert demonstration episodes across *108* household scenes, yielding *176,593* image–text pairs. Tasks were generated/filtered via LLMs over PDDL/TextWorld metadata with paired multimodal demonstrations [155]. *Scoring:* five metrics—*Success Rate (SR)*, *Goal-Condition Success (GCS)*, *Interaction Steps (IS)*, *Language Compliance (LC)*, *Reasoning Disorientation Index (RDI)*—capturing efficiency, action-format validity, and replanning robustness [155].

#### 3) WHAT IT MEASURES

- Structured control-language compliance in action strings [155].
- Replanning robustness under uncertainty/cognitive instability [155].
- Planning efficiency under multimodal observations (IS).

#### 4) WHAT IT LACKS

- *Simulator dependence:* heavy reliance on ALFWorld/ AI2-THOR limits physical realism.
- *Synthetic task bias:* LLM-generated tasks can carry distributional artifacts.
- *Failure patterns in practice:* invalid control strings (LC violations), over/under-exploration inflating IS without completion, and visual-grounding errors driving high RDI [155].

### E. EmbodiedBench

#### 1) WHAT IT IS

A consolidated suite, presented at International Conference on Machine Learning (ICML) 2025, for vision-driven embodied agents based on MLLMs. The suite includes *1,128* testing tasks across four environments (EB-ALFRED, EB-Habitat, EB-Navigation, EB-Manipulation) and six capability-oriented subsets (commonsense, complex instructions, spatial, perception, long-horizon, and basic task solving) [156].

#### 2) COVERAGE AND DESIGN

Unified high-level planning (ALFRED/Habitat) and low-level control (navigation and manipulation) standardize step budgets and stop conditions (e.g., invalid-action caps), and report per-capability scores [156]. *Scoring: task success* aggregated per environment and capability subset under fixed step budgets, with standardized failure taxonomies (e.g., empty plan, too many invalid actions) [156].

#### 3) WHAT IT MEASURES

- Capability breakdowns (commonsense, spatial, perception, long-horizon).
- Sensitivity to visual input vs. text-only ablations.
- Effects of agent design choices (image resolution, visual context length).

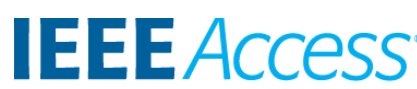

#### 4) WHAT IT LACKS

- *Simulation-centric scope:* limited connection to real-robot execution.
- *Brittleness from upstream environments:* dynamic object indexing/action spaces and inherited quirks from THOR/Habitat.
- *Failure patterns in practice:* sharp drops in low-level manipulation relative to planning (e.g., GPT-4o scores only ∼29% on average); degradation when visual input is ablated [156].

### F. EVALUATING EMBODIED AGENTS: PLANNING AND INTERACTION

Embodied AI places intelligent agents within simulated or physical environments, requiring them to perceive their surroundings and execute actions to achieve the high-level goals specified in natural language [2]. This paradigm fundamentally changes the nature of the evaluation. Success is no longer measured by the accuracy of a question but by the successful completion of a task, such as in assistive technologies such as the AI Guide Dog [157]. The evaluation metrics in embodied AI shift from static correctness to dynamic performance. The key metrics include [117]:

- *Success:* The primary metric, indicating the percentage of tasks that the agent successfully completed.
- *Goal-Condition Success (GCS):* A more fine-grained metric that breaks down the final goal into a set of required state changes (e.g., mug is clean, mug is in coffee maker) and measures the proportion of conditions that are met.
- *Success weighted by Path Length (SPL):* a metric that rewards both success and efficiency, penalizing agents that take unnecessarily long or convoluted paths to complete a task.

These benchmarks test for long-term planning, the ability to decompose a high-level goal into a sequence of executable steps, and the capacity to adapt based on feedback from the environment, a challenge also addressed by frameworks that use LLM explanations to improve apprenticeship learning from human demonstrators [158]. Furthermore, understanding the contributions of different modalities to an agent's decisions is crucial, and frameworks such as MAEA have been developed to compute these attributions and analyze policy behavior [159].

### G. MEASURING THE INTANGIBLE: SOCIAL INTELLIGENCE AND CREATIVITY

Perhaps the most challenging frontier is the evaluation of capabilities that are inherently subjective and deeply rooted in human experiences, such as social intelligence and creativity.

*Social Intelligence:* This involves the ability to perceive and reason about the complex and often unstated dynamics of human social interactions, including emotions, intentions, relationships, and social norms [160]. The *Social-IQ* benchmark was an early attempt to formalize this, providing videos of real-world human interactions and asking multiple-choice questions about the social context (e.g., ''How is the atmosphere of the room?'') [161], [162]. Evaluation in this domain is fraught with several challenges. Human judgment is often the only ground truth and studies have shown that AI and human response patterns can differ significantly. For instance, on Social-IQ, models often default to ''unanswerable'' in ambiguous situations where humans confidently make inferences by leveraging rich visual and auditory cues that are lost in the model [160]. This highlights a key difficulty: evaluating not just the answer, but also the quality and depth of the social reasoning behind it, a challenge that extends to community-centric issues such as detecting violence-provoking speech [163]. Newer frameworks are exploring more complex tasks, such as Inverse Reasoning (deducing motives from actions) to probe these deeper cognitive components [164].

*Creativity:* Measuring the creativity of generative models is another nascent but critical area of research [165]. Researchers have adapted cognitive science frameworks to create quantifiable metrics. A common approach is to automate the scoring of the *Alternative Uses Test (AUT)*, a classic psychological test for divergent thinking where a subject lists as many novel uses as possible for a common object (e.g., a brick) [165]. The AI-generated responses can be scored in the following dimensions:

- *Fluency:* The total number of ideas [165].
- *Flexibility:* The number of different semantic categories the ideas fall into [165].
- *Originality:* The statistical rarity of ideas compared to a baseline corpus [165].

Other approaches draw on the work of cognitive scientist Margaret Boden, who defined three types of creativity—*combinatorial*, *exploratory*, and *transformational*—and attempt to build metrics to assess each [162]. The evaluation of creativity marks a significant departure from objective metrics. The goal is not to measure correctness, but to quantify novelty, surprise, and value—qualities that are inherently tied to human perception and cultural context [162].

The shift toward evaluating these abstract capabilities signifies a maturation in the field's ambitions. The ''cognitive exam'' is moving beyond knowledge and logic to assess the very qualities—interaction, social awareness, and innovation—that are essential for AI to become a truly collaborative partner in the human world.

### H. SYNTHESIS OF LEVEL IV: THE EMBODIED AND CREATIVE FRONTIER

Level IV represents the transition from static reasoning to dynamic agency. As synthesized in Table 26, this frontier pushes the *Skill* axis to its limit, requiring long-horizon planning and social intuition. Interestingly, this level offers a potential solution to the ''Hygiene Crisis'' of Level III: embodied simulators allow for the procedural generation of infinite, unseen testing environments, restoring high *Hygiene* scores. However, this comes at the steepest *Cost* of any era, requiring massive computational resources for physics simulation, and suffers from low *Reliability* due to the inherent subjectivity of evaluating creativity and social dynamics. These emerging challenges in reliability and cost highlight the broader validity threats facing the entire field.

## VII. THREATS TO VALIDITY (AND LIMITS OF THIS SURVEY)

Evaluating multimodal systems entails well-known validity risks [2]. Because this is a *survey* that synthesizes heterogeneous, author-reported results, we do not enforce a single evaluation protocol, re-score models, or audit contamination. Instead, we make the threats explicit and provide *reader guidance* on how to interpret the numbers and comparisons we cite.

### A. HETEROGENEOUS PROTOCOLS AND APPLES-TO-ORANGES COMPARISONS

*Threat.* Reported scores often depend on prompt templates, decoding settings, evaluator choice (human vs. LLM-as-judge), and sometimes even test-time tools [1]. Cross-paper numbers are rarely strictly comparable due to these hidden variables.

*Implications for this survey.* We present results as reported in their original sources. Crucially, we do *not* normalize scores across papers, as differences in compute (e.g., float16 vs. int8), prompt engineering (e.g., 0-shot vs. 5-shot), and evaluator models (e.g., GPT-4 vs. GPT-4o as judge) make post-hoc normalization impossible. Readers should interpret cross-paper comparisons as qualitative trends rather than precise rankings.

### B. BENCHMARK AGING, SATURATION, AND CONTAMINATION

*Threat.* Public test sets age and can overlap with pre-training/tuning corpora; leakage inflates scores and reduces construct validity, as models may rely on memorization rather than reasoning [37], [133].

*Implications for this survey.* While we do not perform independent audits, we interpret reported results through the four-level Audit Recipe defined in Section II, utilizing each benchmark's *Hygiene* rating to contextualize the reliability of public-set performance. Where papers disclose hidden splits or adversarial refreshes, we note it; otherwise, interpret public-set results as upper-bound indicators rather than hygiene-verified measurements.

### C. METRIC CHOICE AND JUDGE DEPENDENCE

*Threat.* Single metrics (e.g., exact match) may misalign with human judgments; LLM-as-judge evaluations can be sensitive to prompts, rubrics, and verbosity, leading to ''length bias'' or self-preference [130], [139].

*Implications for this survey.* We report results using the metrics and judges chosen by the original authors. Where papers note reliance on an LLM judge or a particular rubric, we indicate this to contextualize the score. However, because we do not re-score results under a unified protocol, reported numbers should be treated as *protocol-conditioned* and not assumed directly comparable across papers.

### D. PROMPTING AND DECODING SENSITIVITY

*Threat.* Scores can swing significantly with instruction wording, few-shot exemplars, choice order, temperature/top-$p$, and context length effects [126], [127].

*Implications for this survey.* We did not standardize prompts/decoding across papers. Use trends (large gaps; consistent failures) as stronger evidence than narrow wins. For borderline differences, assume prompt/decoding variance could flip outcomes.

### E. SAMPLING BIAS AND CULTURAL/LINGUISTIC COVERAGE

*Threat.* Datasets over-represent certain domains, languages (often English), and visual cultures (Western-centric); shortcuts and selection artifacts can inflate in-domain performance while masking OOD brittleness [48], [166]. Lessons from multi-view learning suggest that enforcing diversity embeddings and integrating comprehensive classification views are critical for mitigating such biases [167], [168].

*Implications for this survey.* Our benchmark coverage mirrors what is most reported in the literature and is therefore biased toward English, Western, and vision-heavy suites. View gaps we call out (e.g., audio, long-horizon video, non-English OCR) as *coverage limits* of the underlying ecosystem and, by extension, of this survey.

### F. REPRODUCIBILITY AND REPORTING GAPS

*Threat.* Many papers omit seeds, full prompts, or evaluator prompts; some results require proprietary models or data, contributing to a reproducibility crisis in ML research [169].

*Implications for this survey.* We cannot guarantee reproducibility of the cited numbers. Where setup details are missing, we treat results as directional and avoid fine-grained cross-paper rank claims.

#### 1) SCOPE CLARIFICATIONS (WHAT WE DID NOT DO)

We did not (i) re-run models, (ii) harmonize decoding or prompts, (iii) re-score with a uniform judge, (iv) conduct contamination audits, or (v) perform meta-analysis with effect-size normalization. Our contribution is a structured synthesis: *what* each benchmark tests, *how* it scores, and *where* current models tend to fail.

Table 27 outlines the principal validity threats considered in this survey and provides guidance on how to interpret their implications for benchmark evaluation.

## VIII. HUMAN-IN-THE-LOOP "LIVING" BENCHMARKS

Static test sets rapidly saturate and drift out of relevance [3], [13]. Living benchmarks instead run continuous or periodic *human-in-the-loop* collection and curation, often with *adversarial* example generation, and publish versioned releases. These systems better diagnose failure modes, resist overfitting, and reflect evolving user needs [170].

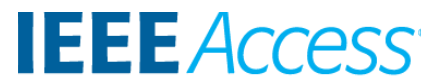

**TABLE 26.** **Framework synthesis: Level IV (Embodiment & Creativity).**

| Dimension | Rating | Justification and Evidence |
|---|---|---|
| *Skill* | ★★★★★ | *Apex Capability.* Requires executing long-horizon plans [117], adapting to physical feedback, and inferring unstated social intent [161]. This exceeds the static Q&A paradigm of all previous levels. |
| *Reliability* | ★★ | *Low (Subjective/Stochastic).* Evaluating creativity (e.g., "Is this story novel?") is inherently subjective [162]. Embodied metrics (Success Rate) suffer from high variance due to simulator physics and partial observability, making reproducibility difficult [42]. |
| *Robustness* | ★★ | *Brittle (Sim-to-Real Gap).* While agents may master a simulator, they often fail to generalize to real-world visual noise or physics (the Sim-to-Real gap) [2]. Adversarial perturbations in planning can cause catastrophic failure in goal execution. |
| *Hygiene* | ★★★★ | *High (Procedural Renaissance).* Unlike static web-scraped datasets, embodied benchmarks (e.g., Habitat, ProcTHOR) can procedurally generate unique, unseen environments for every evaluation run, effectively solving the memorization problem [119]. |
| *Cost* | ★ | *Prohibitive.* The highest computational cost. Running physics simulators for thousands of steps per agent [42] or employing human panels to judge creativity [165] scales poorly compared to static inference. |
| *Fairness* | ★★★ | *Complex.* Simulators allow for controlled diversity (e.g., varying skin tones of avatars), but social intelligence benchmarks rely on human judgment labels that may encode cultural biases regarding "appropriate" behavior [160]. |

**TABLE 27.** **Principal validity threats and how to interpret them in this survey.**

| Threat | Interpretation stance in this survey |
|---|---|
| Heterogeneous protocols | No normalization performed; read cross-paper comparisons as trends only |
| Aging & contamination | We do not audit; flag hidden/refresh splits when authors do; read public-set scores as upper bounds |
| Metric/judge dependence | Retain author-reported metrics; note when results rely on an LLM judge or rubric; treat open-ended scores as *protocol-conditioned* rather than directly comparable |
| Prompt/decoding sensitivity | Emphasize robust gaps/trends over narrow wins; de-emphasize close calls |
| Sampling & coverage bias | Recognize under-covered areas (audio, long-horizon video, non-English OCR) as ecosystem limits |
| Reproducibility gaps | Treat such results as directional; refrain from strong cross-paper rank claims |

### A. DESIGN PATTERNS

*(1) Adversarial data collection.* Annotators (and sometimes models) craft examples that fool strong baselines while remaining valid for humans, yielding harder, more diagnostic items [13], [116], [170]. *(2) Continuous refresh.* Items are added or replaced on a cadence (weekly → semiannual) so scores remain meaningful over time, addressing the *Hygiene* axis [39], [171]. *(3) Human preference arenas.* Crowd-sourced, randomized, pairwise comparisons with Elo-style aggregation capture *user* preferences beyond static metrics, addressing *Alignment* [14]. *(4) Leaderboard hygiene.* Hidden test splits, code-upload evaluation, and anti-leak protocols deter test contamination and tuning-on-test [13], [42], [120]. *(5) Versioned releases.* Public changelogs and semantic versioning (MAJOR.MINOR.PATCH) communicate scope changes, enabling apples-to-apples comparisons across time [172].

### B. REPRESENTATIVE EXEMPLARS

*Dynabench* introduced model+human adversarial rounds across tasks, demonstrating harder, more discriminative datasets that better measure *Robustness* [13]. *ANLI* used iterative, adversarial human-and-model-in-the-loop collection for NLI and showed persistent difficulty for SOTA models [116]. *RealTimeQA* runs weekly evaluations about current events, capturing models' recency and addressing the *Hygiene* crisis by testing on data generated after the training cutoff [39]. Human preference *arenas* (e.g., LMSYS Chatbot Arena) continuously collect head-to-head human judgments (scaling to millions of votes), yielding robust Elo rankings and surfacing instruction-following/verbosity biases that static *Reliability* metrics miss [14]. When static suites saturate or are contaminated, refreshed or contamination-reduced successors (*Massive Multitask Language Understanding (MMLU)-Pro*, *MMLU-CF*) restore *Skill* headroom and *Hygiene* [173], [174]. For the bridge to Level IV, the *Habitat Challenge* evaluates uploaded *agents* in unseen environments annually (code-upload, not prediction-upload), a living pattern now common in embodied benchmarks [42], [119], [120].

### C. RED-TEAMING PIPELINES

Modern pipelines mix human red teamers with *LM-as-red-teamer* generation to expose harms and safety gaps at scale, with iterative attack/defense cycles and category coverage (jailbreaks, leakage, bias, unsafe plans). Public reports and datasets detail methods, scaling behavior, and lessons [151], [170], [175].

### D. MAINTENANCE PLAYBOOK (CADENCE, GOVERNANCE, VERSIONING)

1) *Governance & roles:* Establish a small editorial board (lead, safety chair, data steward). Publish a *datasheet* for each release and maintain *model cards* for reference baselines [176], [177], [178].
2) *Cadence:* Choose a refresh interval matched to domain drift (e.g., weekly for real-time QA; quarterly or

**TABLE 28.** Common "living benchmark" patterns and concrete exemplars.

| Pattern | Instantiation | Key Properties |
|---|---|---|
| Adversarial rounds | Dynabench; ANLI | Human+model-in-the-loop authoring; harder items; iterative releases [13], [116] |
| Continuous refresh | RealTimeQA; LiveBench | Weekly or semiannual re-itemization; recency stress [39], [171] |
| Human arena | LMSYS Chatbot Arena | Randomized pairwise battles; Elo rating system (Elo) rankings at scale [14] |
| Saturation refresh | MMLU-Pro; MMLU-CF | Harder and/or contamination-reduced variants; restored headroom [173], [174] |
| Embodied code upload | Habitat Challenge | Evaluate uploaded *agents* on unseen envs; annual challenge [42], [119], [125] |

semiannual for general reasoning; annual for embodied challenges). Freeze windows precede each release for validation and leakage checks [39], [42], [119], [120], [171].

3) *Adversarial collection:* Run targeted rounds against top models-of-the-day; pay bounties for high-quality failures. Include *challenge sets* per stressor (negation, compositionality, tool-use, safety).
4) *Quality control:* Dual review + arbitration; adversarial validation; item-response analysis (difficulty/discrimination); de-duplication and contamination audits before promotion to test.
5) *Secure evaluation:* Prefer *hidden* or rotating test shards; serve items via Application Programming Interface (API); log access; disallow training-time downloads. For interactive/embodied, require *code upload* and evaluate on *unseen* environments [42], [119], [120].
6) *Versioning & changelogs:* Adopt *Semantic Versioning (SemVer)*: increment MAJOR for test definition changes, MINOR for added items/splits, PATCH for fixes. Publish diffs, migration notes, and deprecated items [172].
7) *Reproducibility checklist:* Require seeded configs, decoding params, prompts/templates, and hardware reports; follow established machine learning (ML) reproducibility checklists [169].
8) *Preference calibration:* For arenas, randomize position/order, normalize verbosity, and regularly re-calibrate raters; publish inter-rater agreement and judge prompts.
9) *Sunset policy:* When saturation occurs (performance > 90% with low variance), demote to training/dev and promote new stressor-focused suites (e.g., *Pro* or *CF* variants) [173], [174].

The recurring patterns used to design and maintain modern "living benchmarks," along with representative examples and their key properties, are summarized in Table 28.

## IX. CONCLUSION: ARE WE BUILDING BETTER MODELS OR BETTER TEST-TAKERS?

The journey of AI evaluation, framed as an evolving cognitive examination, reveals a field in a state of constant, self-correcting motion. We have progressed from the foundational knowledge tests of the recognition era to the complex, diagnostic reasoning exams of today. Yet, this progress raises a crucial, overarching question: are we succeeding in building more intelligent systems, or are we merely becoming more adept at training excellent test-takers? The answer, as this survey suggests, lies in the continuous, adversarial nature of the evaluation process itself.

### A. THE BENCHMARKING DILEMMA

The history of AI benchmarks is rife with a fundamental tension. On one hand, standardized tests are the bedrock of scientific progress, providing the objective, quantifiable measures needed to compare models and track advancements [2]. On the other hand, any static benchmark, no matter how well-designed, represents a fixed target that powerful optimization algorithms will eventually learn to circumvent [13]. This leads to several systemic issues:

- *The "Whac-a-Mole" Problem:* Research has shown that mitigating one specific shortcut or bias in a model can inadvertently cause it to amplify its reliance on another [179]. For example, a model trained to ignore texture bias might become more susceptible to background bias. This suggests that simply patching individual flaws is insufficient; the problem lies deeper in the learning dynamics of the models.
- *The Cycle of Saturation:* As we have seen, benchmarks are being "solved" at an accelerating rate [1]. This forces the community into an expensive and time-consuming cycle of creating new, harder datasets, which themselves will eventually be saturated [180]. This raises concerns that the field is overfitting to the entire evaluation ecosystem, with progress on leaderboards becoming a less reliable proxy for progress on real-world intelligence.
- *Data Contamination:* A pervasive and growing problem is data contamination, where data from evaluation benchmarks inadvertently leaks into the massive, web-scraped datasets used to pre-train foundation models [2], [181]. When a model has been trained on the test questions, its performance is no longer a valid measure of its generalization ability. This undermines the integrity of the entire evaluation paradigm.

These critiques suggest that the current paradigm, while responsible for immense progress, is reaching its limits. The very act of measuring intelligence with a fixed instrument changes the behavior of the systems being measured, often in ways that obscure rather than reveal their true capabilities [182].

### B. OUTLOOK: THE POST-2025 LANDSCAPE

As this survey goes to press in late 2025, the field is witnessing a divergence between generalist saturation and specialist proficiency. While Level III benchmarks such as MMMU are nearing saturation (with *Gemini 3 Pro* is reported to achieve *81%* on MMMU-Pro in Nov 2025 [183]), a new frontier of "Unanswerable Exams" and "World Simulators" is emerging:

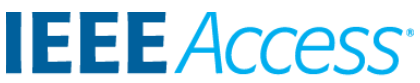

- *The "Last" Exams:* To counter the saturation of MMLU, new "ceiling-test" benchmarks such as *Humanity's Last Exam (HLE)* [184] and *FrontierMath* [185] have been released. Current frontier models (e.g., GPT-4o, Claude 3.5) score below *25%* on HLE and *<2%* on FrontierMath, demonstrating that expert-level *discovery* (as opposed to textbook recall) remains unsolved.
- *System 2 Reasoning Models:* The release of reasoning-specialized models such as *OpenAI o3* has fundamentally shifted the evaluation of abstract logic. Early reports suggest o3 achieves *87.5%* on the ARC-AGI benchmark [186], a task previously considered intractable for LLMs, suggesting that "test-time compute" scaling may break the reliability of static logic puzzles.
- *Generative World Simulators:* As generative video models such as *Sora* and *Veo 2* mature, evaluation is shifting from "describing pixels" to "predicting physics." Emerging research views these models not just as artists, but as *World Simulators* [187], requiring new benchmarks that test causal consistency and physical grounding rather than just visual fidelity.

### C. THE PATH FORWARD: TOWARDS LIVING BENCHMARKS

To resolve the Benchmarking Dilemma, the field must abandon the reliance on static artifacts. As detailed in Section VIII, "Living Benchmarks" offer the necessary structural defense against Goodhart's Law. By implementing *continuous adversarial updates*, we prevent the saturation cycle [13]; by utilizing *dynamic, private test sets* and *code-upload* protocols, we mitigate the contamination crisis [39], [120]; and by integrating *human-in-the-loop arenas*, we align metrics with genuine utility rather than proxy gamification [14]. Three specific directions define this new frontier:

- *Adversarial Human-in-the-Loop Evaluation:* As demonstrated by benchmarks such as AdVQA, using humans to continuously generate novel, model-fooling examples creates a dynamic and ever-evolving test that is much harder to overfit. This approach ensures that evaluation keeps pace with model capabilities, always probing the current edge of their limitations [53].
- *Adaptive Testing from Psychometrics:* Drawing inspiration from human educational testing, researchers are exploring adaptive testing, where the evaluation is tailored to the specific ability level of the model being tested [8]. Instead of a one-size-fits-all exam, an adaptive test can efficiently identify a model's strengths and weaknesses by selecting questions that are most informative for its current performance level, making evaluation faster, fairer, and more precise [188], [189], [190].
- *Evaluation in Interactive Environments:* For capabilities such as long-term planning and social collaboration, the most meaningful evaluation occurs not in a static Q&A format, but within dynamic, interactive environments. Embodied AI benchmarks, where success is measured by task completion in a simulated world, represent a crucial step in this direction [152]. The ultimate test of an agent's intelligence is its ability to reliably and safely achieve goals in a world that reacts to its actions.

The evolution of the AI Cognitive Examination is ultimately a reflection of our own evolving understanding of intelligence. The initial focus on recognition mirrored a belief that intelligence was primarily about knowledge accumulation. The subsequent shift to reasoning reflected a deeper appreciation for logic, compositionality, and causality. The current movement toward evaluating abstract and social intelligence signals an acknowledgment that true intelligence is also embodied, interactive, and creative.

By relentlessly designing new tests that expose the failures of current systems, the research community is engaged in a profound act of scientific inquiry. Each "failed" test provides a clearer picture of what intelligence is not, and in doing so, refines the blueprint for what it could be. The benchmarks are not merely yardsticks; they are the diagnostic tools that reveal the gap between performance and competence, between pattern matching and genuine understanding. However, a critical gap remains: while our tasks now demand reasoning, our metrics largely remain outcome-based proxies. Closing the gap between complex cognitive demands and simplistic evaluation metrics defines the next great challenge for the field. The ongoing effort to close this gap, driven by the design of ever more sophisticated examinations and novel training paradigms like semantic data augmentation [191], is the engine propelling the field beyond the frontiers of perception and toward the horizon of true multimodal reasoning.

## ACKNOWLEDGMENT

The authors made moderate use of OpenAI's ChatGPT (GPT-5) in preparing this manuscript. ChatGPT was used to assist with drafting and refining text in multiple sections, especially sections II-VII. All AI-generated content was critically reviewed, fact-checked, and revised by them to ensure accuracy, originality, and compliance with IEEE standards.

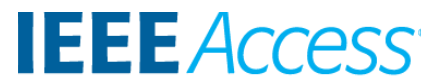

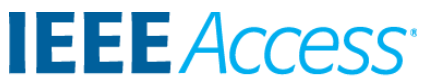

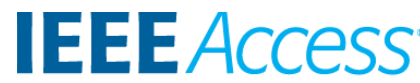

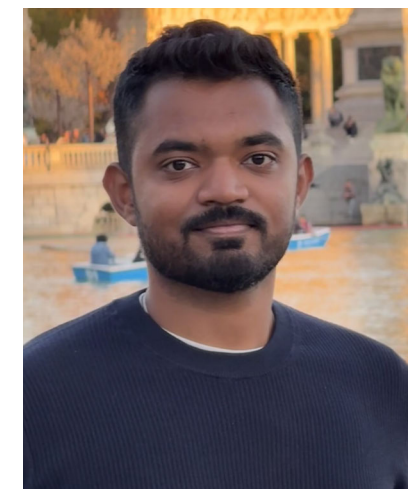

**MAYANK RAVISHANKARA** (Member, IEEE) received the B.E. degree in electronics and communication engineering from R. V. College of Engineering (RVCE), Bengaluru, India, in 2020, and the M.S. degree (Hons.) in information systems management from Carnegie Mellon University (CMU), Pittsburgh, PA, USA, in 2022.

He is currently a Software Engineer at Everlaw, Oakland, CA, USA, and is also an Independent Researcher. He has filed multiple provisional patents in the fields of artificial intelligence and human-computer interaction. His research interests include the development of AI agents, multimodal systems, and privacy-first productivity applications.

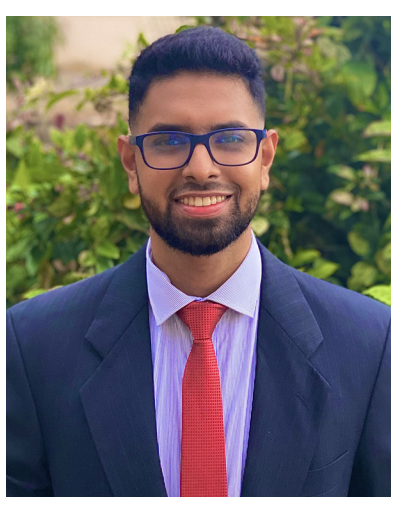

**VARINDRA V. PERSAD MAHARAJ** received the Diploma degree in commercial diving from the Divers Institute of Technology, Seattle, WA, USA, in 2019, and the dual B.S.E. degree in computer science engineering and aerospace engineering from the University of Michigan, Ann Arbor, MI, USA, in 2022.

He began his internship at Amazon, in 2022, on the RISC team as a Software Development Engineer Intern in Sunnyvale, CA, USA. He then continued to work full-time, beginning as a Software Development Engineer, in 2023, again on the RISC team, Sunnyvale. He has research interests in AGI, Robotics, VR/AR, Quantum Computing, and BCI, and is very passionate about entrepreneurship, intending to leverage his technical expertise to found and scale new ventures in tech.